\documentclass[10pt,twocolumn,letterpaper]{article}

\usepackage{cvpr}              % To produce the CAMERA-READY version
\usepackage[dvipsnames]{xcolor}

\definecolor{cvprblue}{rgb}{0.21,0.49,0.74}
\usepackage[pagebackref,breaklinks,colorlinks,citecolor=cvprblue]{hyperref}

\usepackage{times}
\usepackage{epsfig}
\usepackage{graphicx}
\usepackage{amsmath}
\usepackage{amssymb}

\usepackage{multirow}
\usepackage{array}
\usepackage{booktabs}
\usepackage{color}
\usepackage{pifont}
\usepackage{mathtools}
\usepackage{colortbl}
\usepackage{xcolor}
\usepackage{array}
\usepackage{color}
\usepackage{bbding}
\usepackage{makecell}
\usepackage{threeparttable}
\usepackage{float}

\usepackage{relsize}
\def\paperID{3121} % *** Enter the Paper ID here
\def\confName{CVPR}
\def\confYear{2024}

\title{Open-Set Image Tagging with Multi-Grained Text Supervision}
\author{
\vspace{-1.0em}
\\
\vspace{0.2em}
\textbf{Xinyu Huang$^{1,2}$ \quad Yi-Jie Huang$^{2}$ \quad Youcai Zhang$^{2}$ \quad Weiwei Tian$^{4}$ \quad Rui Feng$^{1,4}$}  \\ \vspace{0.2em} \textbf{Yuejie Zhang$^{1}$    \quad Yanchun Xie$^{2}$\quad Yaqian Li$^{2}$ \quad Lei Zhang$^{3}$}  \\
\normalsize
$^{1}$Shanghai Key Lab of Intell. Info. Processing, School of Computer Science, Fudan University \\ 
\normalsize
$^{2}$OPPO Research Institute \quad $^{3}$International Digital Economy Academy~(IDEA) \\ \normalsize $^{4}$Academy for Engineering and Technology, Fudan University}

\begin{document}
\maketitle

% two lines per blah
\newcommand{\blah}{
{blah blah blah blah blah blah blah blah blah blah blah blah blah }
}

% \blah
\newcommand{\bigblah}{\blah \blah \blah \blah \blah}

\newlength\savewidth\newcommand\shline{\noalign{\global\savewidth\arrayrulewidth
  \global\arrayrulewidth 1pt}\hline\noalign{\global\arrayrulewidth\savewidth}}
  
\newcommand{\cmark}{\ding{51}}%

\renewcommand{\thefootnote}{$\dagger$}

\definecolor{xinyu}{rgb}{0.5,0.8,0.7}
\definecolor{pink}{RGB}{219, 41, 145}

\makeatletter
\newcommand{\algorithmfootnote}[2][\footnotesize]{%
  \let\old@algocf@finish\@algocf@finish% Store algorithm finish macro
  \def\@algocf@finish{\old@algocf@finish% Update finish macro to insert "footnote"
    \leavevmode\rlap{\begin{minipage}{\linewidth}
    #1#2
    \end{minipage}}%
  }%
}

\renewcommand{\thefootnote}{\fnsymbol{footnote}}

\definecolor{darkergreen}{RGB}{21, 152, 56}
\definecolor{red2}{RGB}{252, 54, 65}
\newcommand\redp[1]{\textcolor{red2}{(#1)}}
\newcommand\greenp[1]{\textcolor{darkergreen}{(#1)}}

\definecolor{pearDark}{HTML}{2980B9}
\definecolor{pearDarker}{HTML}{1D2DEC}

% \definecolor{pearDark}{RGB}{21, 152, 56}
% \definecolor{pearDarker}{RGB}{21, 152, 56}
\begin{abstract}

In this paper, we introduce the Recognize Anything Plus Model~(RAM++), an open-set image tagging model effectively leveraging multi-grained text supervision.
Previous approaches~(\textit{e.g.,} CLIP) primarily utilize global text supervision paired with images, leading to sub-optimal performance in recognizing multiple individual semantic tags. In contrast, RAM++ seamlessly integrates individual tag supervision with global text supervision, all within a unified alignment framework. This integration not only ensures efficient recognition of predefined tag categories, but also enhances generalization capabilities for diverse open-set categories. Furthermore, RAM++ employs large language models~(LLMs) to convert semantically constrained tag supervision into more expansive tag description supervision, thereby enriching the scope of open-set visual description concepts. Comprehensive evaluations on various image recognition benchmarks demonstrate RAM++ exceeds existing state-of-the-art (SOTA) open-set image tagging models on most aspects. Specifically, for predefined commonly used tag categories, RAM++ showcases 10.2 mAP and 15.4 mAP enhancements over CLIP on OpenImages and ImageNet. For open-set categories beyond predefined, RAM++ records improvements of 5.0 mAP and 6.4 mAP over CLIP and RAM respectively on OpenImages. For diverse human-object interaction phrases, RAM++ achieves 7.8 mAP and 4.7 mAP improvements on the HICO benchmark. Code, datasets and pre-trained models are available at \textcolor[HTML]{3166FF}{\url{https://github.com/xinyu1205/recognize-anything}}.

\end{abstract}
\section{Introduction}
\label{sec:intro}

\begin{figure}[t]
\begin{center}
\includegraphics[width=1.0\linewidth]{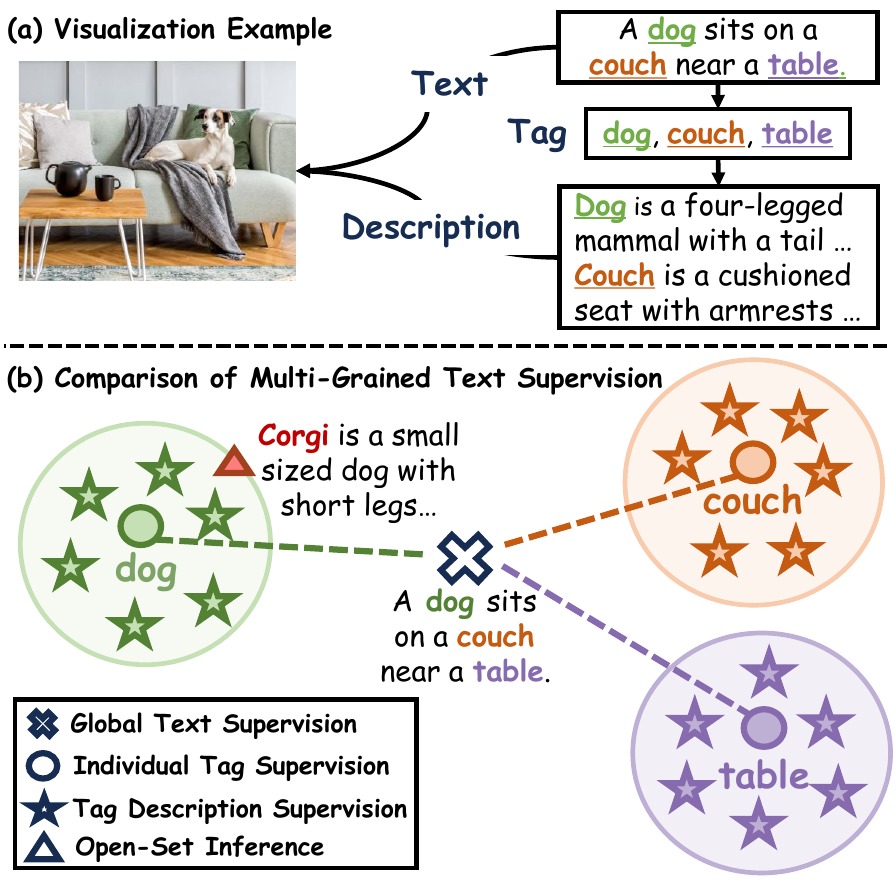}
\end{center}
\vspace{-0.5em}
\caption{\textbf{Illustration of multi-grained text supervision.} (i) Global text supervision entangles multiple semantics, leading to sub-optimal performance in recognizing multiple individual semantic tags. (ii) Our model leverages both individual tag supervision and global text supervision, enhancing tagging capacity on both predefined and open-set categories. (iii) We further convert tag supervision into more expansive tag description supervision via the LLMs, facilitating the recognition of diverse open-set categories with visual concepts.}
\vspace{-0.5em}
\label{fig:intro-1}
\end{figure}

% both global text supervision and individual tag supervision, ensuring 
% integrate individual tag supervision and 
% (ii) Our 
% is sub-optimal in recognizing multiple individual semantic tags, since it 
% entangles multiple semantics, which 
% Prior open-set recognition models primarily utilize global semantic text supervision, leading to sub-optimal performance in recognizing multiple individual semantic tags. Our model leverages multi-grained text supervision, encompassing global semantics and individual tag descriptions, enabling recognition of diverse open-set categories with visual concepts.

Image recognition remains a fundamental research area in computer vision, necessitating machines to output various semantic contents based on the given images. To this end, visual models with text supervision, such as CLIP~\cite{radford2021learning}, ALIGN~\cite{jia2021scaling}, and Florence~\cite{yuan2021florence}, leverage large-scale image-text pairs from the Internet to learn comprehensive visual concepts. These models demonstrate notable open-set recognition in single-label image classification~\cite{deng2009imagenet}, facilitating their application across diverse domain-specific datasets with arbitrary visual concepts~\cite{gu2021open,shen2022k}.

Despite such advances, these models predominantly rely on global text supervision, which directly align global text embeddings with corresponding global visual features. Such supervision is sub-optimal for more complex multi-tag recognition tasks. Due to the global text supervision entangles multiple semantics, the influence of individual tag semantics is significantly weakened. As illustrated in Figure~\ref{fig:intro-1}, the text ``\textit{a dog sits on a touch near a table}" encompasses the concepts of \textit{``dog", ``couch"} and \textit{``table"}. However, its global embedding exhibits partial divergence from these individual semantics.

\begin{figure}[t]
\begin{center}
\includegraphics[width=1.0\linewidth]{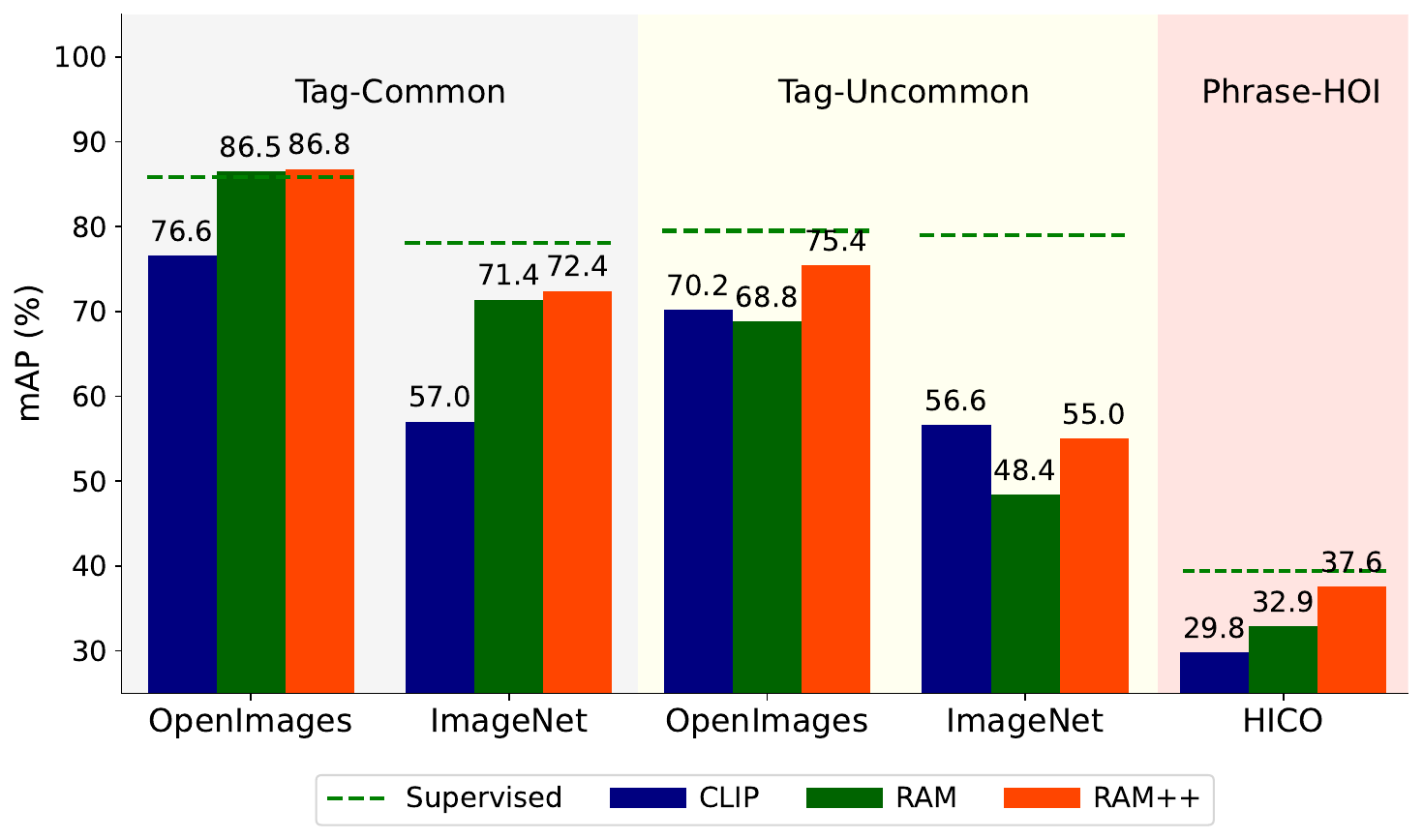}
\end{center}
\vspace{-1.2em}
\caption{\textbf{Comparison of zero-shot image recognition performance on various benchmarks.} Our RAM++ model outperforms existing SOTA open-set image tagging models~(CLIP~\cite{radford2021learning} and RAM~\cite{zhang2023recognize}), in terms of common tag categories of OpenImages and ImageNet, uncommon tag categories of OpenImages and ImageNet, and human-object interaction phrases of HICO.}
\vspace{-0.1em}
\label{fig:intro-2}
\end{figure}

By contrast, image tagging models with individual tag supervision, primarily utilize manually annotated image tags of limited scale~\cite{lin2014microsoft, everingham2015pascal}. Despite recent studies~\cite{huang2023tag2text,zhang2023recognize,huang2022idea} significantly expand the scale of image tags using image-text pairs, image tagging models still fall short in recognizing tag categories beyond their predefined label system. This limitation highlights the constrained semantic generalization capabilities of tag supervision with fixed categories, consequently hindering their broader applicability. For instance, it is challenging to generalize the tag of \textit{``dog''} or \textit{``drinks''} to more specific subcategories such as \textit{``corgi"} or \textit{``Coca Cola''}. Moreover, the numerous phrase categories like \textit{``meteor shower"} further poses this challenge.

To address the aforementioned limitations, our study proposes an open-set image tagging model leveraging multi-grained text supervision, integrating both global text supervision and individual tag supervision. The image tags are automatically parsed from the texts, offering more fine-grained supervision which ensures the competent recognition on predefined tag categories. Simultaneously, the diverse text supervision enables the model to learn a broader range of textual semantics far beyond fixed tag categories, extending generalization capabilities for open-set categories. Specifically, we incorporate image-tag-text triplets within a unified alignment framework. The multi-grained text supervision interacts with visual spatial features through an efficient alignment decoder~\cite{vaswani2017attention}. Compared with other prevalent alignment paradigms, our approach demonstrates superior tagging performance with high efficiency.

Furthermore, considering the insufficient visual concepts of tag supervision, we convert tag supervision into more expansive tag description supervision through large language models~(LLMs)~\cite{chatgpt,gpt4}. LLMs are employed to automatically generate multiple visual descriptions for each tag category. These descriptions are subsequently integrated into tag embedding via a novel automatic re-weighting mechanism, enhancing the relevance with corresponding image features. This approach enriches the scope of visual concepts for the image tagging model, enhancing its capability to incorporate visual descriptions for open-set recognition during inference. For instance, the tag \textit{``corgi"} can be expanded to a more descriptive \textit{``a small-sized dog with short legs ...''}, which aids in determining its presence in images.

\begin{figure*} [t]
\centering
  \includegraphics[width=0.99\linewidth]{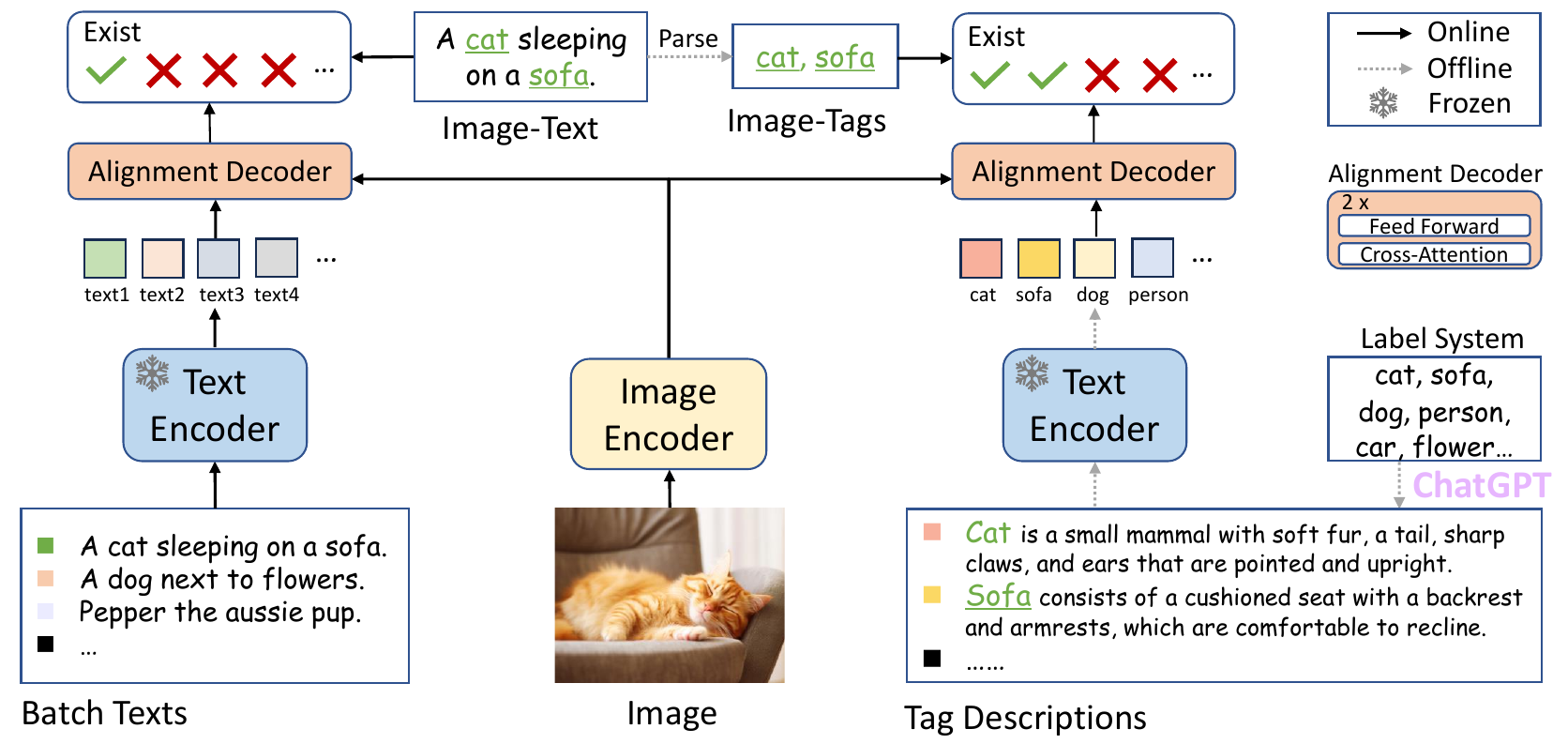}
  \vspace{-0.5em}
  \caption{\textbf{Illustration of RAM++ training framework.} With image-tag-text triplets, RAM++ adopts a shared alignment decoder to align image-text and image-tags simultaneously. The individual tag supervision ensures efficient recognition of predefined tag categories, and the diverse text supervision significantly enhances the open-set tagging abilities. In addition, RAM++ employs a LLM to generate multiple visual descriptions for each category within the label system, thereby enriching the scope of open-set visual concepts. }
  \label{fig:model-architercture}
  \vspace{-0.5em}
\end{figure*}

% This is leveraging by the insight that expand the semantics of tag categories during the the inference phase of the open set tagging model can effectively improve performance~\cite{novack2023chils, menon2022visual, pratt2023does}.

% Specifically, we adopt the LLM to automatically generate diverse visual descriptions for each tag category and synthesize into tag embeddings to align with image features.

% are fixed and 

% recent studies~\cite{novack2023chils, menon2022visual, pratt2023does} have employed the knowledge of LLMs~\cite{brown2020language,chatgpt,gpt4} to boost the inference stage of open-set tagging models. 

% Leveraging this insight, we pioneer the integration of LLM's knowledge into image tagging training. Specifically, we adopt the LLM to automatically generate diverse visual descriptions for each tag category and synthesize into tag embeddings to align with image features. This approach transfer further injects richer semantic information into the model, allowing it to integrate visual description concepts for open-set recognition during inference. In addition, we design an automatic re-weighting mechanism for multiple visual descriptions of a tag category, further enhancing performance without additional complexity consumption.

Consequently, building upon our proposed approaches, we introduce the Recognize Anything Plus Model~(RAM++), an open-set image tagging model with an exceptional capability in recognizing diverse tag categories.
% over 4,500 predefined common categories and diverse open-set categories. 
As depicted in Figure~\ref{fig:intro-2}, RAM++ exceeds existing SOTA open-set image tagging models~(CLIP~\cite{radford2021learning} and RAM~\cite{zhang2023recognize}) across various benchmarks. Notably, RAM++ showcases 10.2 mAP and 15.4 mAP enhancements over CLIP on predefined commonly used categories of OpenImages~\cite{kuznetsova2020open} and ImageNet~\cite{deng2009imagenet}. Moreover, RAM++ also achieves 5.0 mAP and 6.4 mAP improvements over CLIP and RAM on open-set uncommon categories of OpenImages. For diverse human-object interaction phrases, RAM++ achieves 7.8 mAP and 4.7 mAP improvements on HICO~\cite{chao2015hico} against CLIP and RAM, respectively.

% The key contributions of this paper can be summarized as follows:
Our key contributions can be summarized as follows:

\begin{itemize}

\item We integrate the image-tag-text triplets within a unified alignment framework, achieving superior performance on predefined tag categories and augmenting recognition capabilities on open-set categories.

% which not only achieves superior performance on predefined tag categories, but also augments the recognition ability in open-set categories.

\item To the best of our knowledge, our work is the first effort to incorporate LLM's knowledge into image tagging training stage, allowing the model to integrate visual description concepts for open-set category recognition during inference.

\item Evaluations on OpenImages, ImageNet, HICO benchmarks demonstrate that RAM++ exceeds existing SOTA open-set image tagging models on most aspects. Comprehensive experiments provide evidence highlighting the effectiveness of multi-grained text supervision.

\end{itemize}

\section{Related Works}
\label{sec:related}

\vspace{5pt}
\noindent
\textbf{Tag Supervision.} Image tagging, also known as multi-label recognition, involves assigning multiple tags to an image. Traditional methods primarily depend on limited manually annotated datasets~\cite{lin2014microsoft, everingham2015pascal, chua2009nus}, leading to poor generalization capabilities. DualCoop~\cite{sun2022dualcoop} and MKT~\cite{he2023open} employ pretrained vision-language models to boost open-set capabilities, but they are constrained by the scale of training dataset. Tag2Text~\cite{huang2023tag2text} and RAM~\cite{zhang2023recognize} obtain large-scale image tags based on image-text pairs, demonstrating advanced zero-shot capabilities on predefined categories. Nonetheless, all these models rely on tag supervision with closed-set semantic scope, limiting their ability to recognize more diverse range of open-set tag categories. Our RAM++ seamlessly integrate diverse text supervision with tag supervision, effectively enhancing the open-set tagging abilities.

\vspace{5pt}
\noindent
\textbf{Text Supervision.} Visual models with text supervision can recognize open-set categories by aligning visual-linguistic features. Pioneering models like CLIP~\cite{radford2021learning} and ALIGN~\cite{jia2021scaling}, which collect millions of image-text pairs, demonstrate remarkable performance in single-label image classification~\cite{deng2009imagenet}. However, their reliance on global text supervision present challenges in multi-tag tasks of individual semantics~\cite{zhang2023recognize}. Although other studies~(\textit{e.g.,} ALBEF~\cite{li2021align} and BLIP~\cite{li2022blip}) adopt deep visual-linguistic feature fusion, our analysis indicates their limitations of efficiency and capacity in extensive-category tagging tasks. In contrast, RAM++ align multiple texts and individual tags within a unified alignment framework, demonstrating superior tagging performance with high efficiency.

\vspace{5pt}
\noindent
\textbf{Description Supervision.} Several prior works demonstrate the effectiveness of leveraging text-based category descriptions for enhancing image recognition performance. However, all these previous studies rely on external natural language databases such as handcraft~\cite{reed2016learning, he2017fine, huang2021attributes}, Wikipedia~\cite{paz2020zest, elhoseiny2017link} or WordNet~\cite{fellbaum1998wordnet,shen2022k, bujwid2021large,yao2022detclip}. With LLMs~\cite{brown2020language,gpt4} demonstrating powerful knowledge compression capabilities, recent works incorporate LLM's knowledge at the inference stage of CLIP to improve performance~\cite{novack2023chils, pratt2023does,dai2023exploring, liu2023chatgpt, ren2023chatgpt} and interpretability~\cite{menon2022visual}. Different from these approaches, our work pioneers the integration of LLM knowledge into the training process of image tagging, which is natural and effective to enhance the open-set capability of tagging models.

\section{Approaches}

\subsection{Overview Framework}

This section details RAM++, an open-set image tagging model capitalizes from multi-grained text supervision, encompassing both global text supervison and individual tag description supervison. As depicted in Figure~\ref{fig:model-architercture}, the architecture of RAM++ comprises an image encoder, a text encoder, and an alignment decoder. The training data are image-tag-text triplets, comprising image-text pairs and image tags parsed from the texts. During the training process, the input into the model consists of images accompanied with variable batch texts and fixed tag descriptions. Then the model outputs alignment probability scores corresponding to each image-tag/text pair, which are optimized by the alignment loss~\cite{ridnik2021asymmetric}.

% Benefiting from the dual supervision from both texts and tag descriptions, RAM++ is capable of recognizing diverse open-set categories with visual concepts.

% is the images with variable batch texts and fixed tag descriptions, while the output consists of alignment probability scores corresponding to each image-tag/text pair. Benefiting from the dual supervision from both texts and tag descriptions, RAM++ capable of recognizing diverse open-set categories with visual concepts.

% \subsection{Image-Tag-Text Alignment}
\subsection{Multi-Grained Text Alignment}

% Global and Fine-grained Text Supervision}

\vspace{5pt}
\noindent
\textbf{Unified Image-Tag-Text Alignment Paradigm.} With image-tag-text triplets, RAM++ adopts a shared alignment decoder to align image-text and image-tags simultaneously. Figure~\ref{fig:model-architercture} splits the framework into two segments for clarity. The left segment illustrates the process of image-text alignment, where texts from the current training batch are passed through the text encoder to extract global text embeddings. These text embeddings are subsequently aligned with the image features via cross-attention layers in the alignment decoder, where text embedding serves as the Query, and image features as the Key \& Value. Conversely, the right segment emphasizes the process of image tagging, where the image features interact with fixed tag categories using the same text encoder and alignment decoder.

The alignment decoder is a two-layer attention decoder~\cite{vaswani2017attention, liu2021query2label}, each layer comprising a cross-attention layer and a feed-forward layer. This lightweight design ensures the efficiency for image tagging involving extensive categories. Critically, it eliminates the mutual influence between tag embeddings without self-attention layers, thus allowing the model to recognize any quantity of tag categories without affecting performance.
% for image tagging

% the fixed tag categories also pass through the text encoder to extract individual tag embeddings and then interfaces with image features using the shared alignment decoder.

% for image tagging, allow support any quantity tag categories without affecting performance.

% consists of two layers, each consisting of a cross attention layer and a feed forward layer.

% The right segment illustrates the image tagging framework where the image interfaces with fixed category tag embeddings within the alignment decoder. In contrast, the left segment emphasizes our proposed Image-Text Alignment. The texts of the training batch pass through the text encoder to extract individual text embeddings. The text embeddings are subsequently aligned with the image in the shared alignment decoder. By leveraging this approach, our model injects more diverse semantic concepts beyond fixed tag categories.

\begin{figure}[t]
\begin{center}
\includegraphics[width=1.0\linewidth]{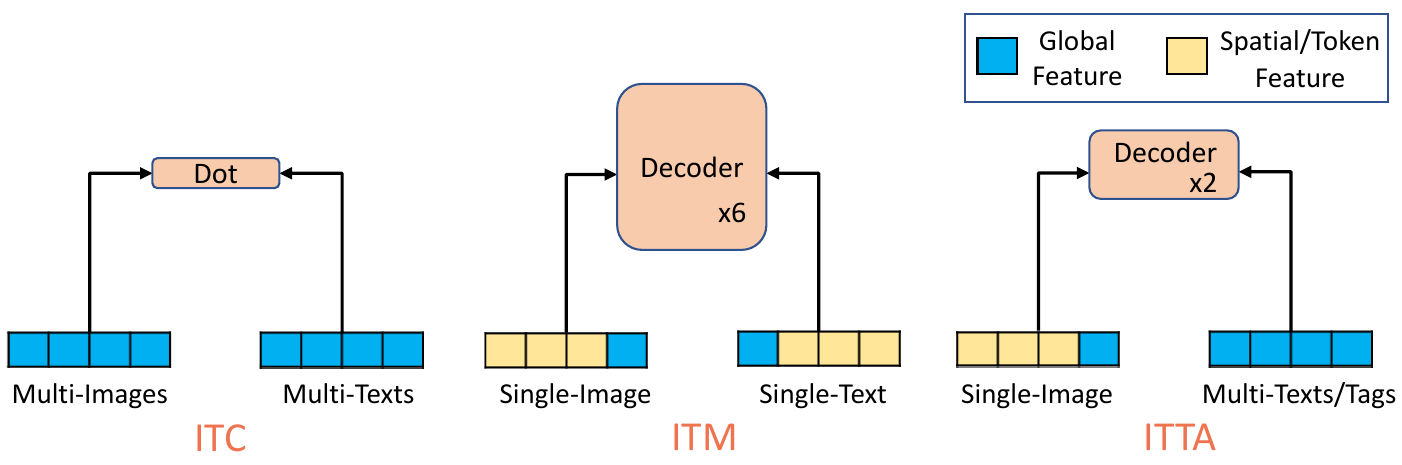}
\end{center}
\vspace{-1.5em}
\caption{\textbf{Comparison of different image-text alignment paradigms:} Image-Text Contrastive Learning~(ITC)~adopted by CLIP~\cite{radford2021learning}, Image-Text Matching~(ITM)~ adopted by BLIP~\cite{li2022blip} and Image-Tag-Text Alignment~(ITTA). Our ITTA unifies image-text alignment with image tagging framework, achieving a balance between efficiency and performance.}
% \vspace{-0.2em}
\label{fig:ITA_compare}
\end{figure}

\vspace{5pt}
\noindent
\textbf{Alignment Paradigm Comparison.} In Figure~\ref{fig:ITA_compare}, we compare our Image-Tag-Text Alignment~(ITTA) with other prevalent alignment paradigms: Image-Text Contrastive Learning~(ITC) adopted by CLIP~\cite{radford2021learning} and ALIGN~\cite{jia2021scaling}, and Image-Text Matching~(ITM) adopted by ALBEF~\cite{li2021align} and BLIP~\cite{li2022blip}. On the one hand, ITC aligns the global features of multiple images and texts simultaneously through dot product with high efficiency. Nonetheless, its reliance on global text supervision with shallow interaction presents challenges for image tagging requiring localized recognition of multiple individual tags. On the other hand, ITM involves in-depth visual-linguistic feature fusions with a deep alignment decoder. However, it only perform one single image-text pair, leading to significant computational costs when aligning the images with multiple texts or tags in both training and inference. Figure~\ref{tab:probability-distribute} demonstrates that both CLIP with ITC and BLIP with ITM fall short in image tagging tasks with sub-optimal performance.

As such, our ITTA addresses these shortcomings by incorporating both global text supervision and individual tag supervision, ensuring robust tagging performance for both predefined and open-set categories. Additional, the adopted efficient alignment decoder utilizes the image spatial feature instead of image global features, taking into account the fact that tags frequently correspond to various image regions. As a result, ITTA establishes a balance between performance and efficiency, capable of aligning the images with thousands of tag categories with high efficiency. For the comparison of inference times across different alignment paradigms, please refer to Figure~\ref{fig:inference_time_compare}.

% Moreover, by deleting the self-attention layers in the alignment decoder, ITA support any quantity of texts or tags without affecting performance.

% ITM can only perform fine-grained interaction to a single image-text pair, leading to significant inference time consumption. 

% in the context of image tagging training, to convert semantically constrained tag supervision into 

% converting semantically constrained tag supervision into 

% we introduce for injecting semantic concepts is LLM-Based Tag description. This approach involves generating distinct descriptions for each tag category, leveraging the capabilities of the LLM in the context of image tagging training.

% \subsection{More Supervision from LLM}

\subsection{LLM-Based Tag Description}

Another innovative approach is LLM-based tag description, which involves leveraging the knowledge of the LLM to convert semantically constrained tag supervision into expansive semantic tag descriptions, thereby enriching the scope of open-set visual concepts that can be described.

\vspace{5pt}
\noindent
\textbf{LLM Prompt Design.} To obtain descriptions for each tag category within the label system, prompt design for LLMs is essential. We anticipate that the tag descriptions generated by LLMs predominantly exhibit two characteristics: {(i)} as diverse as possible to cover a broader range of scenarios;  {(ii)} as relevant as possible to image features for ensuring high relevance.

Drawing inspiration from~\cite{pratt2023does}, we design a total of five LLM prompts for each tag category, as follows: (1) “\textit{Describe concisely what a(n) \{\} looks like}”; (2) “\textit{How can you identify a(n) \{\} concisely?}”; (3) “\textit{What does a(n) \{\} look like concisely?}”; (4) “\textit{What are the identified characteristics of a(n) \{\}}”; (5) “\textit{Please provide a concise description of the visual characteristics of \{\}}”.

\vspace{5pt}
\noindent
\textbf{Tag Description Generation.} Based on the designed LLM prompts, we automatically generate descriptions for each tag category by calling the LLM API. Specifically, we employ the ``GPT-35-turbo" model~\cite{chatgpt}, and set $max\_tokens=77$ which is the same tokenizer length of the text encoder. To promote the diversity of the LLM responses, we set $temperature=0.99$. Consequently, we acquire 10 unique responses for each LLM prompt, amassing a total of 50 tag descriptions per category. Comparison in
Appendix~\ref{app:gpt3-3.5} indicates the superiority of the GPT-3.5 over GPT-3.

% shows that GPT3.5 is better than GPT3.

 % off-the-shelf

\begin{figure}[t]
\begin{center}
\includegraphics[width=0.811\linewidth]{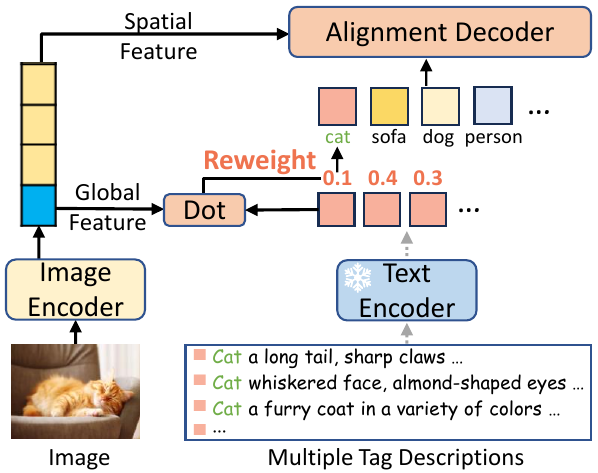}
\end{center}
\vspace{-1.5em}
\caption{\textbf{Automatic re-weighting of multiple tag descriptions.}}
\vspace{-0.5em}
\label{fig:auto_reweight}
\end{figure}

\vspace{5pt}
\noindent
\textbf{Automatic Re-weighting of Multiple Tag Descriptions.} The multiple descriptions of each category requires to be integrated into one tag embedding for image tagging. A straightforward strategy is prompt ensemble, which averages multiple tag descriptions within the textual representation space. This strategy aligns with prevalent works of evaluating on open-set tagging model~\cite{radford2021learning,pratt2023does}. However, the averaged embeddings can be sub-optimal for the training process, due to the ignorance of different similarities between the image and multiple candidate tag descriptions.

To enable selective learning from multiple candidate tag descriptions, we design an automatic re-weighting module for handling multiple tag descriptions, as illustrated in Figure~\ref{fig:auto_reweight}. The probability scores for the i-th tag category are calculated as follows:

\vspace{-1.5em}
\begin{equation}
\scriptsize
% \footnotesize
% \small
\text{Output}_{i} = Decoder [\{V_1,..., V_k\}, \sum_{j=1}^{50} \text{Softmax}(\tau \cdot g_v( V_{\text{global}}) \cdot g_w( \mathbf{d}_{ij})) \cdot \mathbf{d}_{ij}] 
\end{equation}

Where $Decoder$ represents the alignment decoder, $V_{global}$ refers to the image global features and $\{V_1,...,V_k\}$ denotes the image spatial features. The term $\mathbf{d}_{ij}$ signifies the embedding of the j-th tag description. The functions $g_v$ and $g_w$ are projector heads that map inputs into the same dimension, while $\tau$ is a learnable temperature parameter.

% To enable selective learning from multiple candidate tag descriptions, we design an automatic re-weighting scheme for multiple tag descriptions. As illustrated in Figure~\ref{fig:auto_reweight}, for the large volume of categories multiplied by descriptions~(\textit{e.g., $4,500\times50$}), we calculate the dot product between the global image feature and each description, a process with linear time complexity. Then the similarity scores followed by softmax-normalized serve as weights for each description, and ultimately obtain a tag embedding by weighted-sum for each category.

% \begin{equation}
% % \scriptsize
% \footnotesize
% % \small
% \text{Output}_{i} = \text{Decoder}[\{V_1,V_2,..., V_k\}, \sum_{j=1}^{50} \text{Softmax}(\tau \cdot V_{\text{global}} \cdot \mathbf{d}_{ij}) \cdot \mathbf{d}_{ij}] 
% \end{equation}

\subsection{Online/Offline Design}

Our approach also incorporates an online/offline design for different steps, ensuring seamless integration of the image-text alignment and image tagging processes. In the context of image tagging, the number of tag descriptions are fixed but of large volume~(\textit{e.g., $4,500~tag \times 50~des$}). Although extracting embeddings for all tag descriptions is time-consuming, the description embeddings can be pre-processed offline using an off-the-shelf text encoder~\cite{radford2021learning}. In contrast, image-text alignment deals with variable text inputs, where the volume determined by batch size is relatively modest. Therefore, text embeddings can be extracted online for individual batches, circumventing substantial computational cost overhead.

\section{Experiment}

\subsection{Experimental Settings}

\vspace{5pt}
\noindent
\textbf{Training Datasets.} We utilize the same training datasets as that employed by Tag2Text~\cite{huang2023tag2text} and RAM~\cite{zhang2023recognize}. The datasets are based on open-source image-text pair datasets and include two settings: a 4-million (4M) image dataset and a 14-million (14M) image dataset. The 4M setting comprises two human-annotated datasets~(COCO~\cite{lin2014microsoft} and VG~\cite{krishna2017visual}), as well as two web datasets~(SBU Caption~\cite{ordonez2011im2text} and CC-3M~\cite{sharma2018conceptual}). The 14M setting extends the 4M by incorporating CC-12M~\cite{changpinyo2021conceptual}. Our label system includes 4,585 categories that are commonly used in texts. For Tag2Text, the image tags are automatically extracted from their paired texts using a parser~\cite{wu2019unified}. For RAM, both tags and texts are further augmented via an automated data engine~\cite{zhang2023recognize}. We train RAM++ using the RAM datasets, and perform additional validations on the Tag2Text datasets in Appendix~\ref{app:different-datasets}, to substantiate the effectiveness of our proposed methods.

% and also verify using Tag2Text datasets in Appendix~\ref{sec:datasets}, aiming to validate effectiveness of our proposed methods.

% the open-source data and 

% Tag2Text++ and RAM++ using our proposed approaches, aiming to validate their effectiveness.

% \usepackage{multirow}
\begin{table}[t]
\centering
\footnotesize
% \resizebox{\linewidth}{!}{
\begin{tabular}{l|ccc}
\shline
Type                          & Dataset    & \#Images & \#Categories \\ \hline
\multirow{2}{*}{Tag-Common}   & OpenImages & 57,224   & 214          \\
                              & ImageNet   & 5,000    & 492          \\ \hline
\multirow{2}{*}{Tag-Uncommon} & OpenImages & 21,991   & 200          \\
                              & ImageNet   & 5,000    & 508          \\ \hline
Phrase-HOI                    & HICO       & 9,658    & 600          \\ \shline
% Text-Retrieval                & Flickr     & 1,000    & 5,070        \\ \shline
\end{tabular}
% }
\vspace{-0.6em}
\caption{\textbf{The statistics of evaluation benchmarks.}}
\vspace{-0.5em}
\label{tab:benchmarks}
\end{table}

\begin{table*}[]
\resizebox{\linewidth}{!}{
\begin{tabular}{lccccccc}
\shline
\multicolumn{1}{l|}{\multirow{2}{*}{{Methods}}} & \multicolumn{1}{c|}{\multirow{2}{*}{{\begin{tabular}[c]{@{}c@{}}Training\\ \#Images\end{tabular}}}} & \multicolumn{1}{c|}{\multirow{2}{*}{{\begin{tabular}[c]{@{}c@{}}Inference\\ Prompt\end{tabular}}}} & \multicolumn{2}{c|}{{Tag-Common}}                     & \multicolumn{2}{c|}{{Tag-Uncommon}}                   & {Phrase-HOI} \\
\multicolumn{1}{l|}{}                                  & \multicolumn{1}{c|}{}                                                                                       & \multicolumn{1}{c|}{}                                                                                     & {OpenImages} & \multicolumn{1}{c|}{{ImageNet-Multi}} & {OpenImages} & \multicolumn{1}{c|}{{ImageNet-Multi}} & {HICO}       \\ \hline
\multicolumn{8}{l}{\textit{\textbf{Closed-Set Models:}}}                                                                                                                                                                                                                                                                                                                                                                                               \\
\multicolumn{1}{l|}{RelVit~\cite{ma2022relvit}}                            & \multicolumn{1}{c|}{4K}                                                                                     & \multicolumn{1}{c|}{-}                                                                                    & \ding{55}                   & \multicolumn{1}{c|}{\ding{55}}                 & \ding{55}                   & \multicolumn{1}{c|}{\ding{55}}                 & \cellcolor[HTML]{F0FFF0}39.4                \\
\multicolumn{1}{l|}{Swin~\cite{liu2021swin}}                        & \multicolumn{1}{c|}{1.3M}                                                                                     & \multicolumn{1}{c|}{-}                                                                                    & {\ding{55}}                & \multicolumn{1}{c|}{\cellcolor[HTML]{F0FFF0}78.1}                 & \ding{55}                & \multicolumn{1}{c|}{\cellcolor[HTML]{F0FFF0}79.0}                 & \ding{55}                   \\ 
\multicolumn{1}{l|}{ML-Decoder~\cite{ridnik2023ml}}                        & \multicolumn{1}{c|}{9M}                                                                                     & \multicolumn{1}{c|}{-}                                                                                    & \cellcolor[HTML]{F0FFF0}85.8                & \multicolumn{1}{c|}{\ding{55}}                 & \cellcolor[HTML]{F0FFF0}79.5                & \multicolumn{1}{c|}{\ding{55}}                 & \ding{55}                   \\ \hline
\multicolumn{1}{l|}{\multirow{2}{*}{Tag2Text~\cite{huang2023tag2text}}}         & \multicolumn{1}{c|}{4M}                                                                                     & \multicolumn{1}{c|}{-}                                                                                    & 82.9                & \multicolumn{1}{c|}{\ding{55}}                 & \ding{55}                   & \multicolumn{1}{c|}{\ding{55}}                 & \ding{55}                   \\
\multicolumn{1}{l|}{}                                  & \multicolumn{1}{c|}{14M}                                                                                    & \multicolumn{1}{c|}{-}                                                                                    & 83.4                & \multicolumn{1}{c|}{\ding{55}}                 & \ding{55}                   & \multicolumn{1}{c|}{\ding{55}}                 & \ding{55}                   \\ \hline \hline
\multicolumn{8}{l}{\textit{\textbf{Open-Set Models:}}}                                                                                                                                                                                                                                                                                                                                                                                                 \\
\multicolumn{1}{l|}{MKT$^*$~\cite{he2023open}}                              & \multicolumn{1}{c|}{162K}                                                                                   & \multicolumn{1}{c|}{Hand-Written}                                                                         &   77.8              & \multicolumn{1}{c|}{54.7}              &       63.5          & \multicolumn{1}{c|}{45.2}              & 25.5                \\ \hline
\multicolumn{1}{l|}{BLIP$_{ITC}$~\cite{li2022blip}}                              & \multicolumn{1}{c|}{129M}                                                                                   & \multicolumn{1}{c|}{Hand-Written}                                                                         & 75.7                & \multicolumn{1}{c|}{56.2}              & 61.1                & \multicolumn{1}{c|}{36.4}              & 33.5                \\

\multicolumn{1}{l|}{\textcolor{lightgray}{BLIP$_{ITM}$~\cite{li2022blip}}}                              & \multicolumn{1}{c|}{\textcolor{lightgray}{129M}}                                                                                   & \multicolumn{1}{c|}{\textcolor{lightgray}{Hand-Written}}                                                                         &  \textcolor{lightgray}{71.7}               & \multicolumn{1}{c|}{ \textcolor{lightgray}{50.8}}              &     \textcolor{lightgray}{62.9}            & \multicolumn{1}{c|}{ \textcolor{lightgray}{37.9}}              &    \textcolor{lightgray}{38.0}             \\ \hline

\multicolumn{1}{l|}{DiHT~\cite{radenovic2023filtering}}                              & \multicolumn{1}{c|}{438M}                                                                                   & \multicolumn{1}{c|}{Hand-Written}                                                                         &   71.3             & \multicolumn{1}{c|}{67.7}              &   62.4             & \multicolumn{1}{c|}{\textbf{66.8}}              &     36.7           \\ \hline

\multicolumn{1}{l|}{\multirow{2}{*}{CLIP~\cite{radford2021learning}}}             & \multicolumn{1}{c|}{400M}                                                                                   & \multicolumn{1}{c|}{Hand-Written}                                                                         & 73.6                & \multicolumn{1}{c|}{56.6}              & 66.2                & \multicolumn{1}{c|}{58.6}              & 26.8                \\
\multicolumn{1}{l|}{}                                  & \multicolumn{1}{c|}{400M}                                                                                   & \multicolumn{1}{c|}{LLM Tag Des}                                                                          & 76.6                & \multicolumn{1}{c|}{57.0}              & 70.2                & \multicolumn{1}{c|}{56.6}              & 29.8                \\ \hline
\multicolumn{1}{l|}{\multirow{3}{*}{RAM$^*$~\cite{zhang2023recognize}}}              & \multicolumn{1}{c|}{4M}                                                                                     & \multicolumn{1}{c|}{Hand-Written}                                                                         & 86.0                & \multicolumn{1}{c|}{70.2}              & 66.7                & \multicolumn{1}{c|}{47.3}              & 32.8                \\
\multicolumn{1}{l|}{}                                  & \multicolumn{1}{c|}{14M}                                                                                    & \multicolumn{1}{c|}{Hand-Written}                                                                         & 86.5                & \multicolumn{1}{c|}{71.4}              & 68.8                & \multicolumn{1}{c|}{48.4}              & 32.9                \\
\multicolumn{1}{l|}{}                                  & \multicolumn{1}{c|}{14M}                                                                                    & \multicolumn{1}{c|}{LLM Tag Des}                                                                          & 82.2                & \multicolumn{1}{c|}{62.8}              & 65.9                & \multicolumn{1}{c|}{43.2}              & 29.6                \\ \hline
\multicolumn{1}{l|}{\multirow{2}{*}{RAM++$^*$}}            & \multicolumn{1}{c|}{4M}                                                                                     & \multicolumn{1}{c|}{LLM Tag Des}                                                                          & 86.5                & \multicolumn{1}{c|}{71.6}              & 73.9                & \multicolumn{1}{c|}{51.3}              & \textbf{37.8}                \\
\multicolumn{1}{l|}{}                                  & \multicolumn{1}{c|}{14M}                                                                                    & \multicolumn{1}{c|}{LLM Tag Des}                                                                          & \textbf{86.6}                & \multicolumn{1}{c|}{\textbf{72.4}}              & \textbf{75.4}                & \multicolumn{1}{c|}{55.0}              & 37.7                \\ \shline
\end{tabular}
}
\vspace{-0.6em}
\caption{\textbf{Zero-shot performance comparison of SOTA open-set image tagging models on mAP.} \colorbox[rgb]{0.93,1.0,0.93}{Green} refers to fully supervised learning with vertical domain training datasets. Inference prompt refers to the category prompt during model inference, \textit{e.g.,} Hand-Written: ``\textit{A photo of a cat}"; LLM Tag Description: ``\textit{Cat is a small general with soft fur ...}". BLIP$_{ITM}$ requires more than 1000$\times$ inference time of CLIP and RAM++ in recognizing thousands of tag categories~(see Figure~\ref{fig:inference_time_compare}). $^*$ indicates the models leveraging the off-the-shelf CLIP. }
\vspace{-0.2em}
\label{tab:main-table}
\end{table*}

\vspace{5pt}
\noindent
\textbf{Implementation Details.} We employ the Swin$_{Base}$~\cite{liu2021swin} pre-trained on ImageNet~\cite{deng2009imagenet} as the image encoder, and select base-scale models across other comparative methods for fair comparison. We leverage the off-the-shelf text encoder from CLIP~\cite{radford2021learning} to extract text and tag description embeddings. We adopt the robust alignment loss function of ASL~\cite{ridnik2021asymmetric} for both image-text alignment and image tagging. The comparison of different alignment loss functions is available in Appendix~\ref{app:different-loss-functions}. Following~\cite{li2021align,li2022blip,huang2023tag2text,zhang2023recognize}, our model further fine-tunes on the COCO dataset after pre-trianing to augment its performance. Benefiting from the fast convergence characteristic, the 4M and 14M versions of RAM++ necessitate only 1 and 3 days respectively for training, using 8 A100 GPUs.

\vspace{5pt}
\noindent
\textbf{Evaluation Benchmarks.}  We employ mean Average Precision~(mAP) as the evaluation metric, which is well-established for evaluating multi-tag recognition performance~\cite{ridnik2021asymmetric,liu2021query2label,ridnik2023ml, zhang2023recognize}. Additional metrics, including F1 scores, precision, and recall, are provided in Appendix~\ref{app:add-eval-metric}.

We assess the image tagging capabilities on various out-of-domain evaluation benchmarks. Specifically, we utilize the widely used benchmarks OpenImages~\cite{kuznetsova2020open} and ImageNet~\cite{deng2009imagenet}. Given that ImageNet is single-labeled and has missing labels in its test set~\cite{yun2021re,beyer2020we}, we resort to ImageNet-Multi~\cite{beyer2020we}, where each image in the test set possesses multiple labels for a more comprehensive annotation. The categories of these benchmarks are categorized into ``common" and ``uncommon" categories based on the inclusion within the RAM++ label system. For more evaluations on the phrase categories, we resort to the HICO~\cite{chao2015hico} benchmark, a prevalent standard on human object interactions~(HOI). HICO encompasses 80 object categories, 177 action categories, resulting in a total of 600 ``human-act-object" phrase combinations. The statistics of the evaluation benchmarks are presented in Table~\ref{tab:benchmarks}. It is worth noting that for RAM and RAM++, apart from Tag-Common which are considered as predefined categories, all other benchmarks refer to unseen categories in an open-set configuration.

% as it provides a comprehensive assessment for multi-tag recognition capability~\cite{ridnik2021asymmetric,liu2021query2label, zhang2023recognize}

\begin{table*}[t]
\captionsetup{type=figure}
\centering
\resizebox{\linewidth}{!}{
\begin{tabular}{cccc}
\hline
\multicolumn{2}{c}{\textbf{Text Supervision}} & \textbf{Tag Supervision} & \textbf{Text + Des Supervision} \\ \hline
\textbf{CLIP}              & \textbf{BLIP$_{ITM}$}             & \textbf{RAM}             & \textbf{RAM++}                \\ \hline
\includegraphics[width=0.3\textwidth]{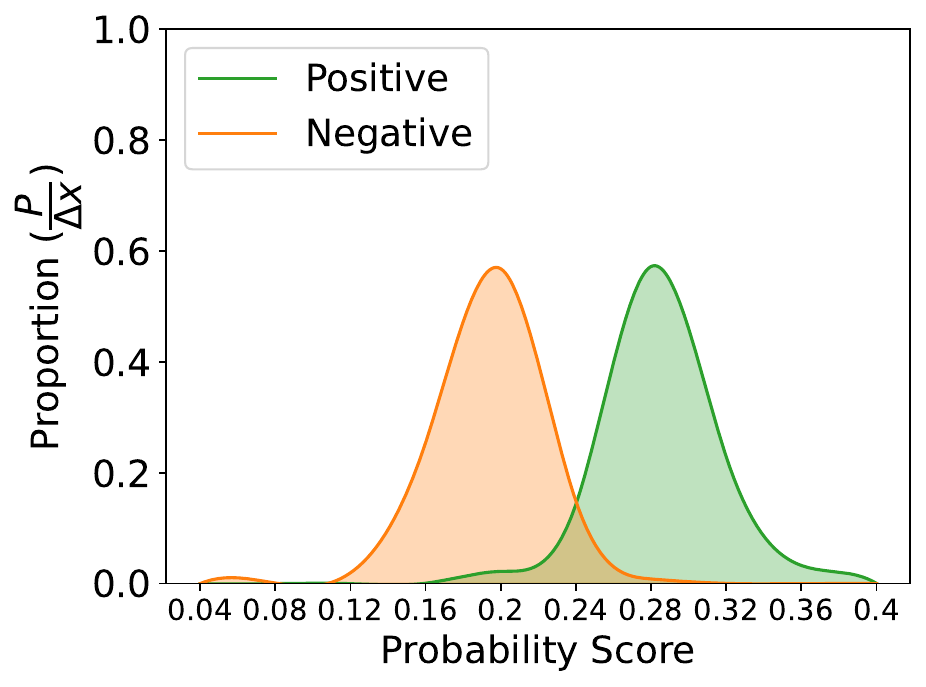}            & \includegraphics[width=0.3\textwidth]{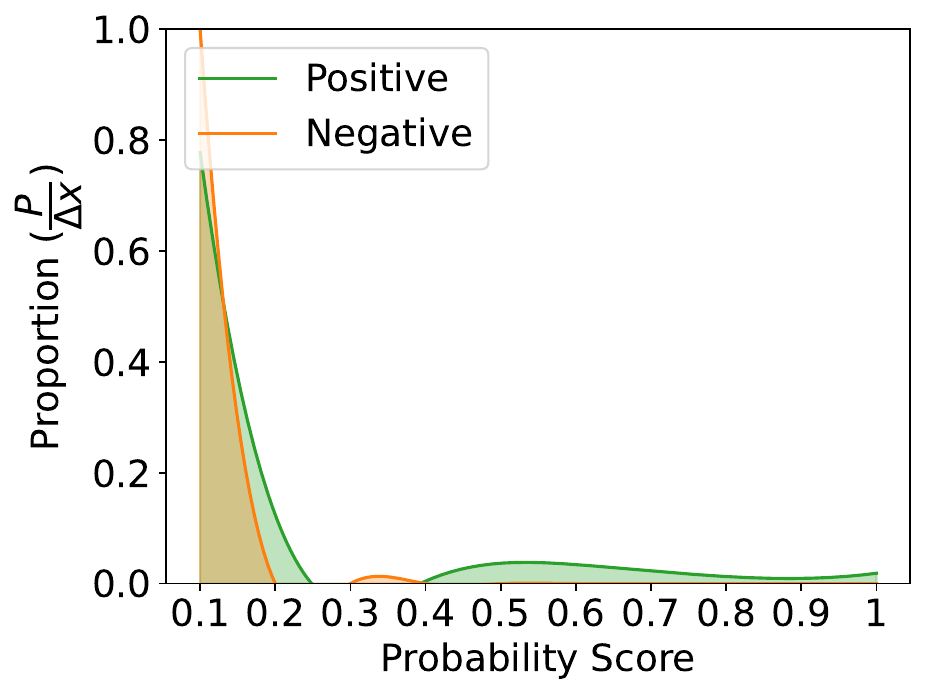}          & \includegraphics[width=0.3\textwidth]{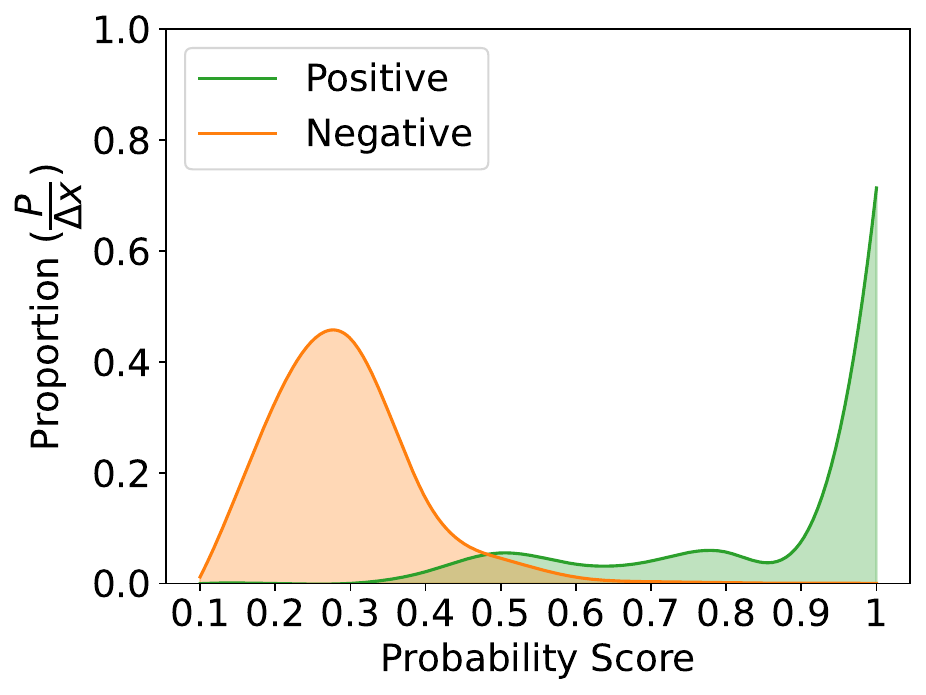}         & \includegraphics[width=0.3\textwidth]{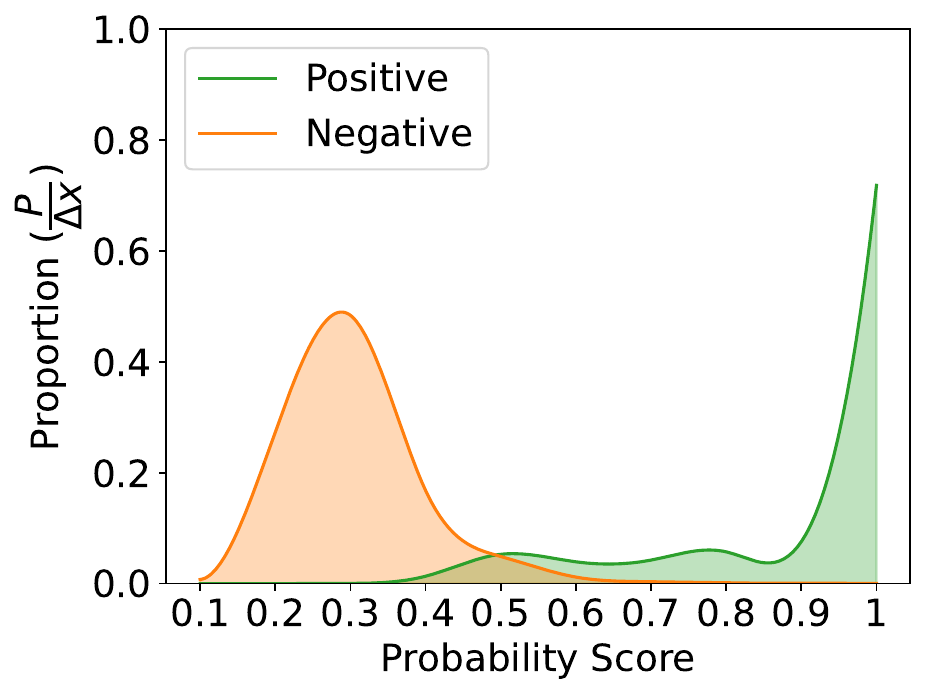}              \\ 
\multicolumn{4}{c}{Common Tag Categories}                                     \\ \hline
\includegraphics[width=0.3\textwidth]{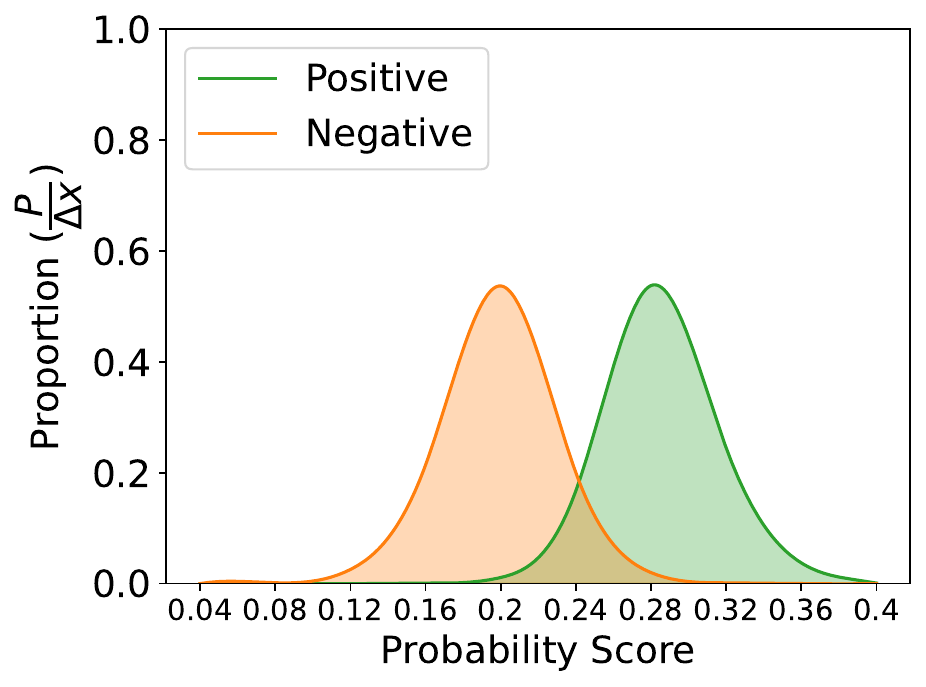}            & \includegraphics[width=0.3\textwidth]{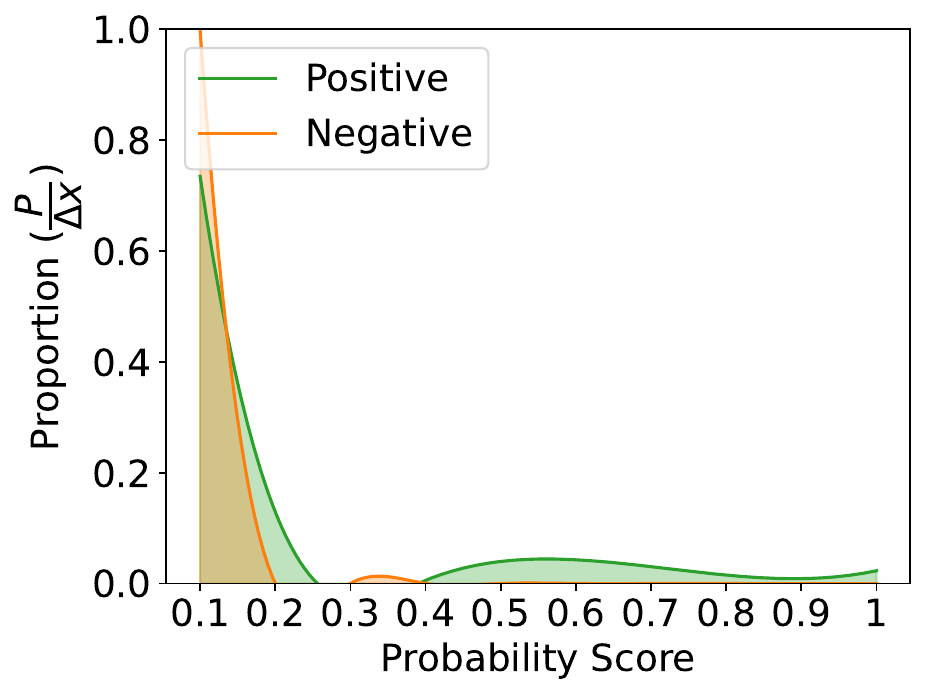}          & \includegraphics[width=0.3\textwidth]{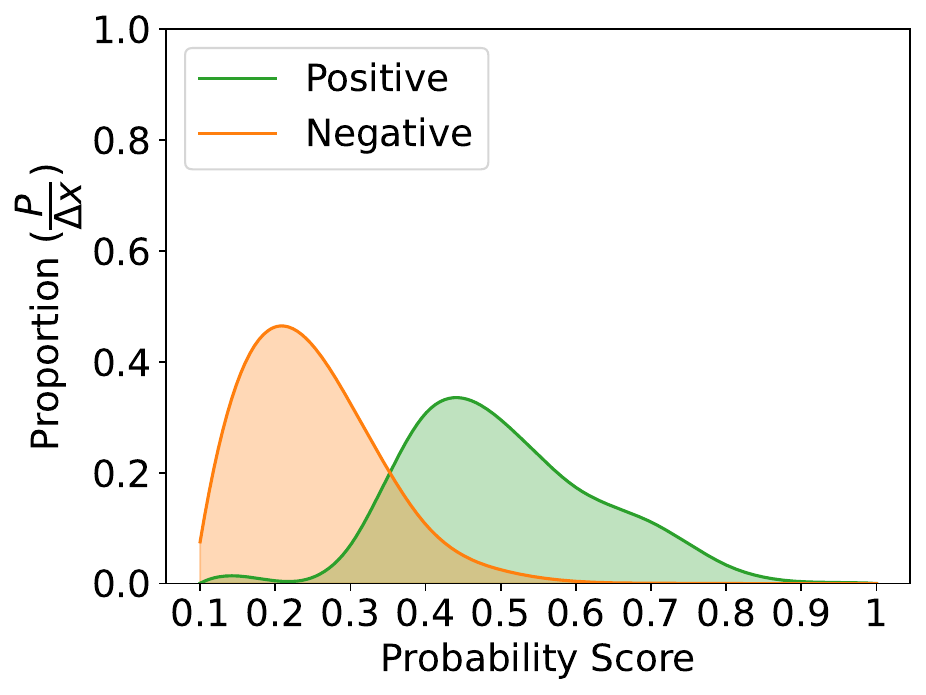}         & \includegraphics[width=0.3\textwidth]{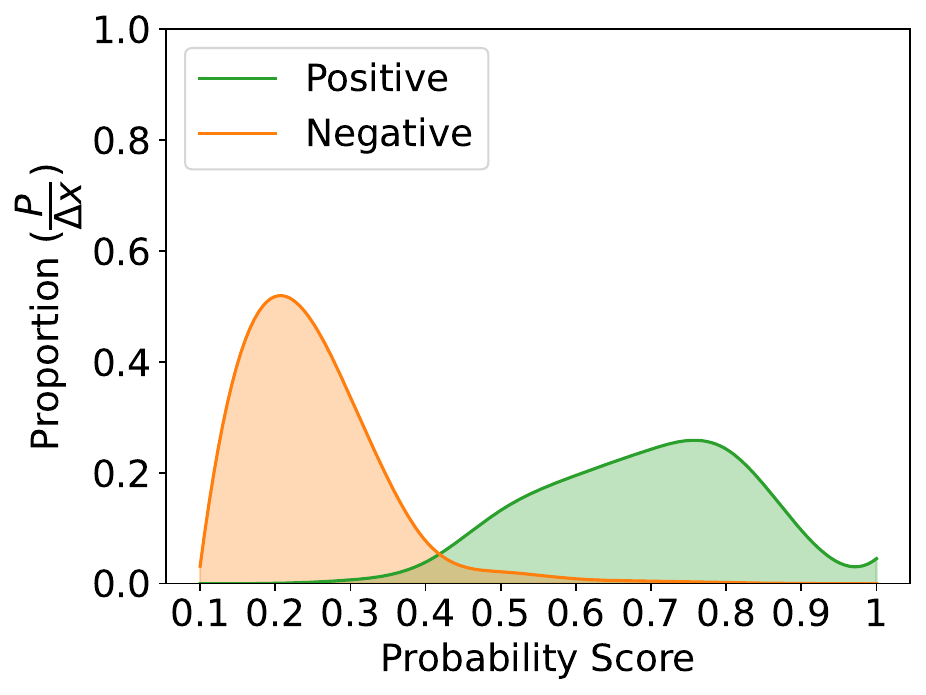}              \\ 
\multicolumn{4}{c}{Uncommon Tag Categories}                                     \\ \hline
\end{tabular}
}
\vspace{-0.8em}
\caption{\textbf{Distribution of probability scores for positive and negative tags on the OpenImages benchmark.} On the one hand, text-supervised models, such as CLIP and BLIP, exhibit challenges in predicting high probability scores for positive tags, leading to sub-optimal performance for multi-tag recognition. On the other hand, the tag-supervised model RAM falls short in recognizing open-set categories. As such, our RAM++, which leverages both text and tag description supervision, demonstrates robust performance across both predefined common and open-set uncommon tag categories.}
\vspace{-0.3em}
\label{tab:probability-distribute}
\end{table*}

\subsection{Comparison with State-of-the-Arts}

\renewcommand{\thefootnote}{$\dagger$}

\vspace{5pt}
\noindent
\textbf{Quantitative Results.} Table~\ref{tab:main-table} presents the zero-shot \footnote{Zero-shot refers to the model does not utilize the training dataset of the corresponding vertical domain.} performance comparison between RAM++ and SOTA open-set image tagging models. On the one hand, text-supervised models such as BLIP and CLIP, exhibit sub-optimal performance across both common and uncommon categories on multi-tag recognition. On the other hand, the tag-supervised model RAM notably boosts performance on common categories, but falls short on uncommon categories compared to CLIP. Moreover, the performance of CLIP can be significantly enhanced when utilizing the LLM tag descriptions for inference, which is consistent with the findings of~\cite{pratt2023does}. Conversely, RAM does not benefit from LLM tag descriptions, indicating its limited open-set generalization potential due to the constrained semantics of tag supervision.

Our RAM++ model, which utilizes both text supervision and tag description supervision, establishes a new SOTA zero-shot performance across various benchmarks. Specifically, RAM++ outperforms CLIP by 10.0 mAP and 15.4 mAP on the common categories of OpenImages and ImageNet, respectively. In terms of open-set categories, RAM++ significantly outperforms RAM on both Tag-Uncommon and Phrase-HOI, underscoring the effectiveness of our approach. Remarkably, RAM++ achieves an improvement of 6.6 mAP and 5.2 mAP over RAM and CLIP on OpenImages-uncommon, and 8.0 mAP and 4.9 mAP over RAM and CLIP on HICO, respectively. 

Despite RAM++ slightly behind CLIP on the uncommon categories of ImageNet, we attribute to that the 14M dataset scale of RAM++ is inadequate for covering these rare categories. It is noteworthy that the data expansion from 4M to 14M for RAM++ result in a 3.7 mAP performance improvement on ImageNet-Uncommon. We contend that further scaling up the training dataset could potentiate the open-set recognition efficacy of RAM++.

\begin{table*}[]
\resizebox{\linewidth}{!}{
\renewcommand{\arraystretch}{1.34}
\footnotesize
\begin{tabular}{c|cccc|c|cc|cc|c}
\shline
\multirow{2}{*}{Case} & \multirow{2}{*}{\begin{tabular}[c]{@{}c@{}}Text\\ Supervision\end{tabular}} & \multirow{2}{*}{\begin{tabular}[c]{@{}c@{}}Tag\\ Supervision\end{tabular}} & \multirow{2}{*}{\begin{tabular}[c]{@{}c@{}}Tag Description\\ Supervision\end{tabular}} & \multirow{2}{*}{\begin{tabular}[c]{@{}c@{}}Automatic\\ Weighting\end{tabular}} & \multirow{2}{*}{\begin{tabular}[c]{@{}c@{}}Inference\\ Prompt\end{tabular}} & \multicolumn{2}{c|}{Tag-Common} & \multicolumn{2}{c|}{Tag-Uncommon} & Phrase-HOI \\
                      &                                                                             &                                                                            &                                                                                    &                                                                                &                                                                             & OpenImages      & ImageNet      & OpenImages       & ImageNet       & HICO       \\ \hline
(a)                   & \ding{51}                                                                           & \multicolumn{1}{l}{}                                                       & \multicolumn{1}{l}{}                                                               & \multicolumn{1}{l|}{}                                                          & Hand-Written                                                                & 77.4            & 47.0          & 69.6             & 38.5           & 31.9       \\
(b)                   &                                                                             & \ding{51}                                                                          &                                                                                    &                                                                                & Hand-Written                                                                & 86.0            & 70.2          & 66.7             & 47.3           & 32.8       \\
(c)                   & \ding{51}                                                                           & \ding{51}                                                                          &                                                                                    &                                                                                & Hand-Written                                                                & 86.5            & 71.5          & 70.5             & 49.9           & 35.5       \\ \hline
(d)                   & \ding{51}                                                                           & \ding{51}                                                                          &                                                                                    &                                                                                & LLM Tag Des                                                                 & 83.1            & 67.2          & 71.6             & 47.7           & 35.6       \\
(e)                   & \ding{51}                                                                           &                                                                            & \ding{51}                                                                                  &                                                                                & LLM Tag Des                                                                 & 86.5            & 71.3          & 73.4             & 50.8           & 37.2       \\
(f)                   & \ding{51}                                                                           &                                                                            & \ding{51}                                                                                  & \ding{51}                                                                              & LLM Tag Des                                                                 & 86.6            & 71.6          & 73.9             & 51.3           & 37.8       \\ \shline
\end{tabular}
}
\vspace{-0.6em}
\caption{\textbf{Ablation study of multi-grained text supervision} on various image tagging benchmarks.}
\vspace{-0.3em}
\label{tab:ablation-study}
\end{table*}

\vspace{5pt}
\noindent
\textbf{Distribution of Probability Scores.} In Figure~\ref{tab:probability-distribute}, we analyze the distribution of probability scores for positive and negative tags across various models on the OpenImages benchmark. An effective model should clearly distinguish between positive and negative tags. Notably, RAM++, wtih dual supervision from texts and tag descriptions, demonstrates robust performance on both predefined and open-set tag categories. 

Besides, we acknowledge the value of investigating the reasons behind the score distributions of different alignment paradigms, which we leave as future work. As an illustration, we consider the contrastive loss in CLIP may leading to its scores around 0.2. And the suboptimal distribution of the ITM model can be attributed to the insufficient utilization of negative samples during training.

\begin{figure} [t]
\centering
\vspace{-0.8em}
  \includegraphics[width=0.9\linewidth]{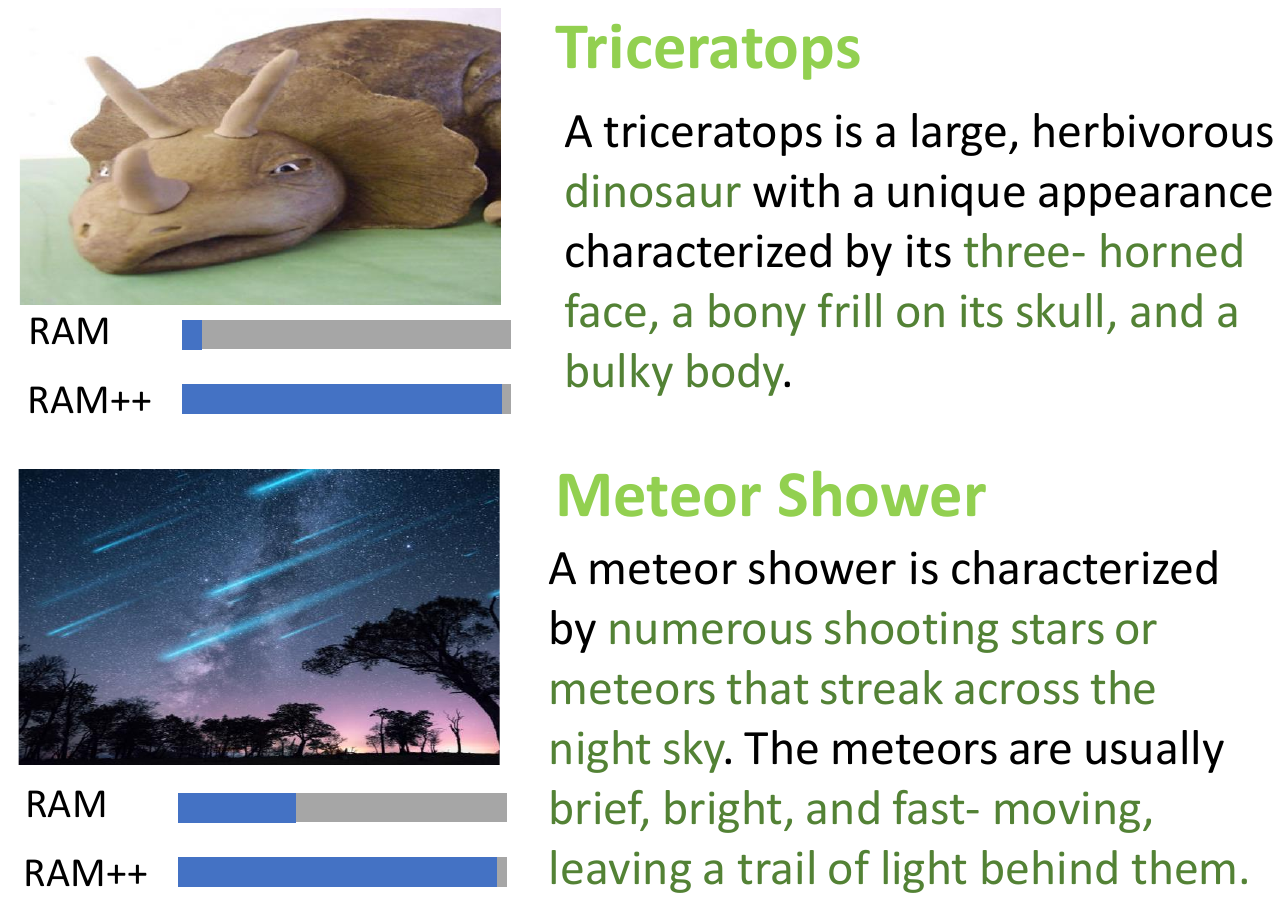}
\vspace{-0.5em}
  \caption{\textbf{Visual comparison of probability scores} from RAM and RAM++ for open-set category recognition based on tag descriptions. The descriptions are those assigned the highest weight by the RAM++ re-weighting module.}
  \label{fig:visual-examples}
  % \vspace{-0.8em}
\end{figure}

Quantitative results of prediction probability comparison between RAM and RAM++ are provided in Figure~\ref{fig:visual-examples}. The descriptions depicted in the figure represent those with high weight in automatic re-weighting. RAM++ demonstrates a significant improvement in prediction probabilities on open-set categories.

% consider it is valuable to explore and analyze the reasons behind the scores distribution of different alignment paradigms which we leave it as future work. For example, we consider  that the contrastive loss results in a score of around 0.2 for CLIP, while ITM has difficulty separating positive and negative samples due to its complexity in training, as it has not seen enough negative samples.

% The results in Figure~\ref{tab:probability-distribute} reveal that text-supervised models struggle to predict high probability scores for positive tags, while tag-supervised model falls short in recognizing open-set categories. In contrast, our RAM++ model, with dual supervision from text and tag descriptions, demonstrates robust performance across both predefined and open-set tag categories.

\subsection{Analysis of Multi-Grained Supervision}

\vspace{5pt}
\noindent
\textbf{Evaluation on Multi-Grained Text Supervision.} We conduct a comprehensive ablation study in Table~\ref{tab:ablation-study} to evaluate the impact of multi-grained text supervision. Case (a) and (b) refer to the two segments of Figure~\ref{fig:model-architercture}, which leverage solely text supervision and tag supervision through the alignment decoder. Text supervision maintains consistent performance across various benchmarks, whereas tag supervision enhances outcomes in common categories.

Case (c) demonstrates the superiority of integrating image-text alignment with image tagging, significantly enhances the model's capability to recognize open-set categories, evidenced by a 3.8 mAP and 2.7 mAP improvement on OpenImages-Uncommon and HICO. This approach, in contrast to the tag-supervised RAM model referenced in Table~\ref{tab:main-table}, avoids a sharp decline in performance when utilizing LLM tag descriptions as the inference prompts, suggesting an enhanced semantic concepts by text supervision.

Case (e) underscores the effectiveness of incorporating LLM tag descriptions in the training stage. When also employing tag descriptions for open-set categories evaluation, our model records the 2.9 and 1.7 mAP improvements on OpenImage-Uncommon and HICO. Such results indicates that expanding the semantically restricted tag supervision into a wide range of descriptive concepts during both training and inference stage, can substantially yield benefits for open-set tagging recognition.

Building on this foundation, case (f) reveals the automatic re-weighting of multiple tag descriptions further enhance the model's capabilities. In Section~\ref{app:weight-compare}, we showcase our re-weighting module achieves more significant improvements with more specific and diverse tag descriptions.

% We also provide quantitative results of prediction probability comparison between RAM and RAM++ in Figure~\ref{fig:visual-examples}. The descriptions depicted in this figure represent those with high weight in automatic re-weighting. Notably, RAM++ demonstrates a significant improvement in prediction probabilities on open-set categories.

% We consider that enhancing the diversity of tag descriptions can lead to more pronounced improvements, which we leave it as future work.

\begin{figure}[]
\begin{center}
\includegraphics[width=1.0\linewidth]{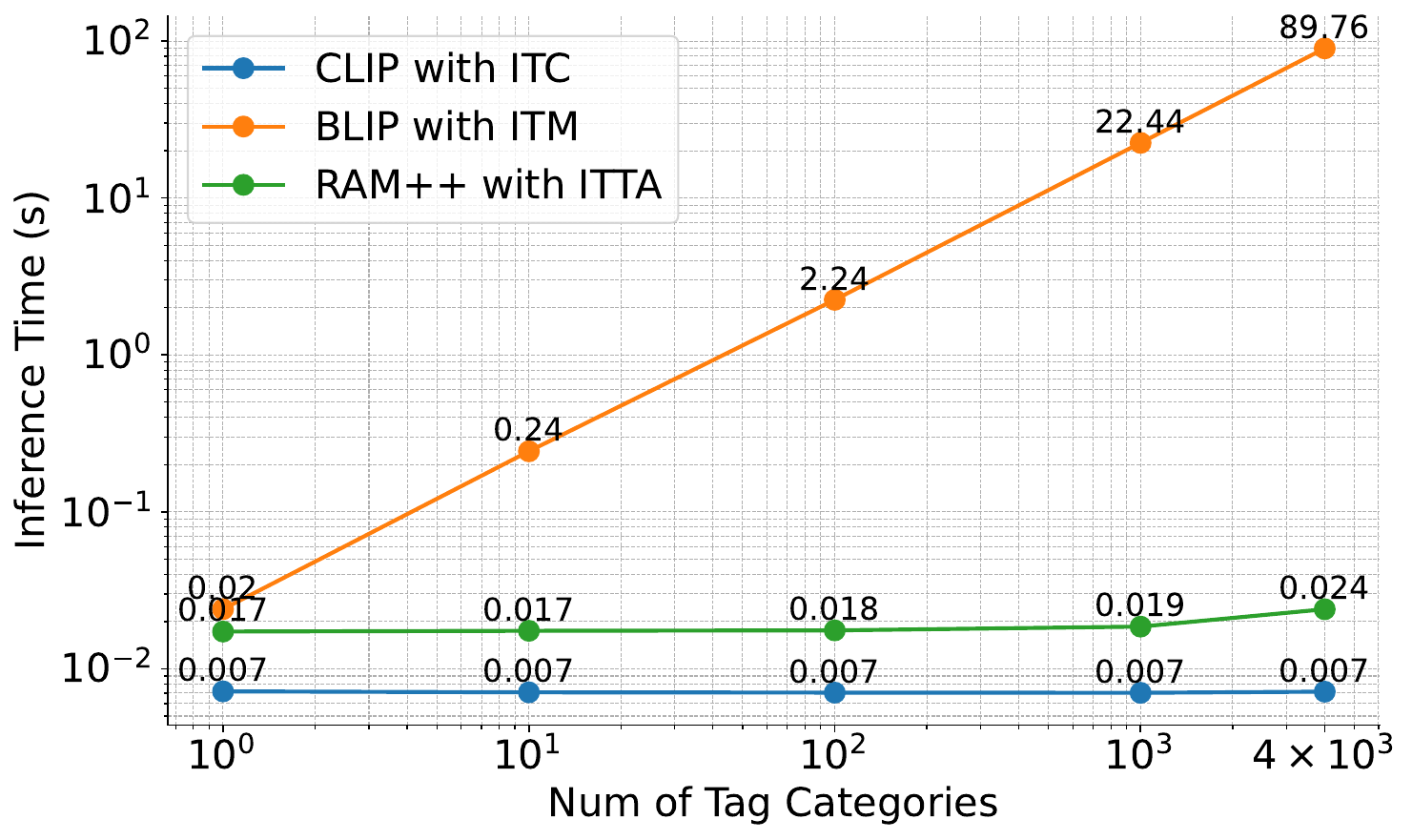}
\end{center}
\vspace{-1.5em}
\caption{\textbf{Inference time comparison between different alignment paradigms} for an image with the number of tag categories increasing.}
\label{fig:inference_time_compare}
\end{figure}

% ITC adopted by CLIP~\cite{radford2021learning}, ITM adopted by BLIP~\cite{li2022blip}, and ITTA adopted by RAM++.

\vspace{5pt}
\noindent
\textbf{Inference Time Comparison.} 
Figure~\ref{fig:inference_time_compare} presents the comparison of inference time consumption across three alignment paradigms with the number of tag categories increasing. This comparison utilizes the average inference time calculated over 1,000 iterations, conducted on an A100 GPU. The figure obviously reveals that inference time for ITM models, which align with a single image-text pair, increases exponentially with the augmentation of categories. This trend poses challenges for the model when processing a large array of tag categories. In contrast, the ITC and ITTA models maintain high inference efficiency, even with a large increase on tag categories. For instance, in the scenario of recognizing 4,000 categories, the ITM model requires 86.76 seconds, whereas the ITC and ITTA models necessitate only 0.024 seconds and 0.007 seconds. 
% respectively.

% \usepackage{multirow}
\begin{table}[]
\begin{center}
% \resizebox{\linewidth}{!}{
\footnotesize
\begin{tabular}{cc|ccc}
\shline
\multirow{2}{*}{\begin{tabular}[c]{@{}c@{}}Image\\ Feature\end{tabular}} & \multirow{2}{*}{\begin{tabular}[c]{@{}c@{}}Feature\\ Fusion\end{tabular}} & \multicolumn{2}{c}{OpenImages-} & \multirow{2}{*}{HICO} \\
                                                                         &                                                                              & Common        & Uncommon       &                       \\ \hline
Global                                                                   & Dot Product                                                                  & 85.0          & 68.9           & 34.5                  \\
Spatial                                                                  & Align Decoder                                                                      & 85.5          & 73.8           & 37.8                  \\ \shline
\end{tabular}
% }
\vspace{-0.5em}
\caption{\textbf{Performance comparison of  image features with different granularities.} }
\vspace{-1.8em}
\label{tab:itc-ita-compare}
\end{center}
\end{table}

\vspace{5pt}
\noindent
\textbf{Comparison of Image Features with different granularities.} Table~\ref{tab:main-table} demonstrates that RAM++ with ITTA consistently outperforms CLIP with ITC across various benchmarks. To further compare image features of different granularity, we conduct the evaluation of image spatial features with the alignment decoder, against image global features with dot product, under the same training dataset comprising image-tag-text triplets. As indicated in Table~\ref{tab:itc-ita-compare}, image spatial features consistently outperform global features, particularly on OpenImages-Uncommon and HICO benchmarks of open-set categories. These results highlight the significance of our ITTA, seamlessly integrates image-text alignment and image tagging within the fine-grained alignment decoder framework.

% \usepackage{multirow}
\begin{table}[t]
\centering
\footnotesize
\begin{tabular}{c|c|cc}
\shline
\multirow{2}{*}{\begin{tabular}[c]{@{}c@{}}Description\\ Type\end{tabular}} & \multirow{2}{*}{\begin{tabular}[c]{@{}c@{}}Multiple\\ Description\end{tabular}} & \multicolumn{2}{c}{ImageNet-} \\
                                                                            &                                                                                 & Common       & Uncommon      \\ \hline
\multirow{2}{*}{Basic}                                                      & Ensemble                                                                        & 65.3         & 46.0          \\
                                                                            & Reweight                                                                        & 65.5         & 46.5          \\ \hline
\multirow{2}{*}{Specific}                                                   & Ensemble                                                                        & 60.1         & 25.7          \\
                                                                            & Reweight                                                                        & 62.7         & 31.9          \\ \shline
\end{tabular}
\vspace{-0.3em}
\caption{\textbf{Performance comparison of different integrated methods for multiple tag descriptions.} }
\vspace{-0.2em}
\label{tab:weight-compare}
\end{table}

\vspace{5pt}
\noindent
\textbf{More Specific and Diverse Descriptions.} We observe that the diversity of LLM descriptions, controlled by temperature, is mainly limited to rephrasing rather than offering true semantic variety. To further validate the effectiveness of our proposed automatic re-weighting of multiple tag descriptions, we attempt to employ more specific and diverse tag descriptions. Specifically, we design the LLM prompt of ``\textit{Describe 50 different possible appearances of what a(n) \{\} looks like}" to generate descriptions. Table~\ref{tab:weight-compare} illustrates that our automatic re-weighting module achieves more significant improvements with more specific and diverse tag descriptions, due to the proposed freedom to selectively learn from mutually different texts. However, there is also a significant decline on the quality of these descriptions, leading to much lower overall performance than the basic version. 

\label{app:weight-compare}

\section{Conclusion}

This paper introduces RAM++, an open-set image tagging model with robust generalization capabilities. By leveraging multi-grained text supervision, RAM++ achieves exceptional performance across various open-set categories. Comprehensive evaluations demonstrate that RAM++ exceeds existing SOTA models on most aspects. Given the revolution in natural language process by LLMs, RAM++ highlights that integrating the knowledge of natural language can significantly empower visual models. We hope our efforts can provide some inspiration for other works.

{
    \small
    \bibliographystyle{ieeenat_fullname}
    \bibliography{main}
}

\clearpage
% \setcounter{page}{1}
% \maketitlesupplementary

\appendix

\section{More Implementation Details}
\label{app:more-detail}

Our models are uniformly pre-trained 5 epochs with a batch size of 720, followed by a fine-tuning process through an additional epoch on the higher-quality COCO dataset~\cite{lin2014microsoft}. The optimizer is the AdamW~\citep{loshchilov2017decoupled} with a weight decay of 0.05. During the pre-training stage, the input images are resized to $224\times224$. The learning rate is warmed-up to $1e^{-4}$ over the first 3,000 iterations, and then follows linear decay with a rate of 0.9. In the fine-tuning stage, the input images size increase to $384\times384$ and the learning rate is set to $5e^{-6}$. Following~\cite{zhang2023recognize, he2023open}, we employ the CLIP image encoder paired with the frozen text encoder to distill image feature, making full use of its original image text alignment properties.

\section{Comparison with Open-Set Localization Models}

This section provides a comparative analysis between RAM++ and other SOTA open-set localization models~(detection~\cite{liu2023grounding} and segmentation~\cite{xu2023open}). The SAM~\cite{kirillov2023segment} model is not included in the comparison due to its lack of recognition capabilitiesa. Table~\ref{tab:detect-segment-recognize} illustrates the zero-shot recognition performance of different models on ADE20K~\cite{zhou2017scene}~(including 143 categories). Notably, RAM++ demonstrates significant advantages on both precision and recall metrics.

More importantly, the efficiency of these localization models exhibits a highly correlation with the quantity of categories need to be recognized. Specifically, they can effectively locate the corresponding objects when provided with the correct image tags. However, their recognition and localization performance markedly decline when provided with a large number of indeterminate categories.

In contrast, RAM++ maintains the robust recognition ability across thousands of categories with high accuracy. This distinctive capability enables RAM++ can significantly empower localization models to develop a strong visual semantic analysis pipeline.

\begin{table}[h]
\centering
\footnotesize
\begin{tabular}{lcc}
\shline
\multicolumn{1}{l|}{\multirow{2}{*}{Methods}} & \multicolumn{2}{c}{ADE20k} \\
\multicolumn{1}{l|}{}                         & Precision     & Recall     \\ \hline
\multicolumn{3}{l}{\textit{Open-Set Detection Model:}}                                    \\
\multicolumn{1}{l|}{Grounding-DINO~\cite{liu2023grounding}}           & 35.6          & 26.0       \\ \hline
\multicolumn{3}{l}{\textit{Open-Set Segmentation Model:}}                                 \\
\multicolumn{1}{l|}{ODISE~\cite{xu2023open}}                    & 48.2          & 50.3       \\ \hline
\multicolumn{3}{l}{\textit{Open-Set Recognition Models:}}                                  \\
\multicolumn{1}{l|}{CLIP~\cite{radford2021learning}}                     & 31.0          & 5.5        \\
\multicolumn{1}{l|}{RAM++}                    & 54.0          & 52.4       \\ \shline
\end{tabular}
\caption{\textbf{Tagging performance comparison of RAM++ with other SOTA open-set localization models.}}
\label{tab:detect-segment-recognize}
\end{table}

% a noteworthy distinction lies in the recognition performance of these localization models is highly correlated to the quantity of categories that need to recognize. Specifically, when the correct tags of an image is given to the localization models, the localization models can effectively locate the corresponding objects. However, when given a large number of uncertain categories, the localization models cannot accurately recognize and locate.

\section{Evaluation on Image-Text Retrieval}

We extend our evaluation on image-text retrieval task to assess the model's alignment ability with fine-grained text. Specifically, we focus on text-to-image retrieval performance of Flickr30K~\cite{plummer2015flickr30k}, given its prominent application in practical scenarios. As depicted in Table~\ref{tab:retrieval}, RAM substantially underperforms compared to CLIP, which further substantiate the limited generalization ability of RAM for open-set semantics. Our RAM++, which employs the same dataset as RAM, even outperforms CLIP on both R@5 and R@10 metrics, demonstrating the effectiveness of our proposed approaches. In addition, although BLIP achieves the best performance among zero-shot models, it relies on ITC+ITM, resulting in a considerable inference time — remarkably longer than both CLIP and RAM++ by several magnitudes.

% \usepackage{multirow}
\begin{table}[h]
\begin{center}
% \resizebox{\linewidth}{!}{
\footnotesize
\begin{tabular}{lcccc}
\shline
\multicolumn{1}{l|}{\multirow{2}{*}{Methods}} & \multicolumn{1}{c|}{\multirow{2}{*}{\begin{tabular}[c]{@{}c@{}}Time/query\\ (ms)\end{tabular}}} & \multicolumn{3}{l}{Text-Retrieval (Flickr30K)} \\
\multicolumn{1}{l|}{}                         & \multicolumn{1}{c|}{}                      & R@1          & R@5          & R@10         \\ \hline
\multicolumn{5}{l}{\textit{Fine-tuned Models:}}                                                                                                           \\
\multicolumn{1}{l|}{UNITER~\cite{chen2020uniter}}                   & \multicolumn{1}{c|}{-}                     & 75.6         & 94.1         & 96.8         \\
\multicolumn{1}{l|}{ERNIE-ViL~\cite{yu2021ernie}}                & \multicolumn{1}{c|}{-}                     & 76.7         & 93.6         & 96.4         \\
\multicolumn{1}{l|}{VILLA~\cite{gan2020large}}                    & \multicolumn{1}{c|}{-}                     & 76.3         & 94.2         & 96.8         \\ \hline \hline
\multicolumn{5}{l}{\textit{Zero-Shot Models:}}                                                                                                           \\
\multicolumn{1}{l|}{CLIP~\cite{radford2021learning}}                     & \multicolumn{1}{c|}{$\sim$0.6}                    & \textbf{68.7}         & 90.6         & 95.2         \\
\multicolumn{1}{l|}{RAM~\cite{zhang2023recognize}}                      & \multicolumn{1}{c|}{$\sim$3.1}                   & 45.9         & 75.9         & 84.6         \\
\multicolumn{1}{l|}{RAM++~(Ours)}                    & \multicolumn{1}{c|}{$\sim$3.1}                   & 66.8         & \textbf{92.0}         & \textbf{95.8}         \\ \hline \hline
% \multicolumn{1}{l|}{BLIP$_{ITC}$}                  & \multicolumn{1}{l|}{12s}                   & 79.8         & 95.0         & 97.7         \\
\multicolumn{1}{l|}{BLIP~\cite{li2022blip}}              & \multicolumn{1}{c|}{$\sim$402.4}               & 85.0         & 96.8         & 98.6         \\ \shline
\end{tabular}
% }
\caption{\textbf{Text to image retrieval performance comparison.}}
\label{tab:retrieval}
\end{center}
\end{table}

\section{Additional Evaluation Metrics}
\label{app:add-eval-metric}

In Table~\ref{tab:add-metric}, we present additional evaluation metric results, including F1 score, precision and recall. We manually adjust the threshold of different models to ensure comparability across evaluations. The results demonstrate that our RAM++ exceeds other open-set image tagging models in both predefined and open-set categories, further highlights the robust tagging capabilities of RAM++.

% \usepackage{multirow}
\begin{table}[h]
\footnotesize
\centering
\begin{tabular}{l|ccc|ccc}
\shline
\multirow{2}{*}{Methods} & \multicolumn{3}{c}{OpenImages-Common}         & \multicolumn{3}{c}{OpenImages-Uncommon}       \\
                         & F1            & Precision     & Recall        & F1            & Precision     & Recall        \\ \hline
BLIP                     & 64.8          & 78.6          & 55.1          & 53.9          & 54.7          & 53.1          \\
CLIP                     & 63.0          & 77.9          & 52.9          & 63.8          & 55.8          & 73.7          \\
RAM                      & \textbf{77.6} & 79.5          & \textbf{75.9} & 54.0          & 53.8          & 54.3          \\
RAM++                    & \textbf{77.6} & \textbf{79.9} & 75.4          & \textbf{64.8} & \textbf{56.3} & \textbf{76.2} \\ \shline
\end{tabular}
\caption{Zero-shot performance comparison with SOTA open-set image tagging models in various metrics.}
\label{tab:add-metric}
\end{table}

\section{GPT3 vs. GPT3.5.}
\label{app:gpt3-3.5}

% \vspace{5pt}
% \noindent
% \textbf{GPT3 vs. GPT3.5.} 

In Table~\ref{tab:llm-compare}, we compare the performance impact of using different LLMs to generate tag descriptions for RAM++ (LLM with consistent training and testing). Evaluation results suggest that GPT-3.5 offers superior performance compared to GPT-3, due to its enhanced accuracy and diversity in responses.

In addition to the LLMs, we also attempt to utilize WordNet descriptions~\cite{fellbaum1998wordnet}. However, their contribution to performance was minimal, due to  WordNet only provides one description or even no description for each category.

% \usepackage{multirow}
\begin{table}[h]
\centering
\footnotesize
\begin{tabular}{l|cc}
\shline
\multirow{2}{*}{LLM} & \multicolumn{2}{c}{Tag-Uncommon} \\
                     & OpenImages       & ImageNet      \\ \hline
GPT-3                & 72.9             & 55.4          \\
GPT-3.5              & 73.8             & 55.5          \\ \shline
\end{tabular}
\caption{\textbf{Performance comparison of different LLMs applied in RAM++. }}
\label{tab:llm-compare}
\end{table}

\section{Validation on Different Training Datasets} 
\label{app:different-datasets}

We further validate our approaches on the 4M training dataset of Tag2Text. Tag2Text fully utilizes open-source image-text pairs. RAM further augments both tags and texts via an automated data engine. As shown in Table~\ref{tab:validate-tag2text}, RAM++ demonstrates notable improvements across various benchmarks on both training datasets, highlighting the efficacy of our approaches.

% both Tag2Text++ and RAM++ based on our methods demonstrate notable improvements across various benchmarks, highlighting the efficacy of our approaches.

% \usepackage{multirow}
\begin{table}[h]
\begin{center}
% \resizebox{\linewidth}{!}{
\relsize{-2.5}
\begin{tabular}{l|c|ccc}
\shline
\multirow{2}{*}{Training Dataset}                                                           & \multirow{2}{*}{Method} & Tag-Common & Tag-Uncommon & Phrase-HOI \\
                                                                                   &                         & OpenImages & OpenImages   & HICO       \\ \hline
\multirow{3}{*}{Image-Text Pairs}                                                        & Tag2Text                & 82.9       & \ding{55}            & \ding{55}          \\
                                                                                   & RAM              & 83.1       & 63.2         & 28.4       \\ 
                                                                                    & RAM++              & 83.5       & 70.4        & 35.6       \\\hline
\multirow{2}{*}{\begin{tabular}[c]{@{}l@{}}Image-Text Pairs\\ +Data Engine\end{tabular}} & RAM                     & 86.0       & 66.7         & 32.8       \\
                                                                                   & RAM++                   & 86.5       & 73.9         & 37.8       \\ \shline
\end{tabular}
% }
\caption{\textbf{Approaches validation on different training datasets.}}
\label{tab:validate-tag2text}
\end{center}
\end{table}

\section{Alignment Loss Function Comparison}
\label{app:different-loss-functions}

\vspace{5pt}
\noindent
\textbf{Image-Text Alignment Loss Function.} In Table~\ref{tab:ita-loss} and Table~\ref{tab:tagging-loss}, we compare different alignment loss functions for image-text alignment and image tagging, including the Cross Entropy~(CE) function employed by CLIP, and other robust tagging loss functions~(BCE, ASL~\cite{ridnik2021asymmetric}, Hill~\cite{zhang2021simple}, SPLC~\cite{zhang2021simple}). The results indicate that ASL outperforms other loss functions, which alleviates the potential missing labels and imbalance between positive and negative samples.

% \usepackage{multirow}
\begin{table}[h]
\footnotesize
\centering
\begin{tabular}{c|cc}
\shline
\multirow{2}{*}{\begin{tabular}[c]{@{}c@{}}ITA\\ Loss\end{tabular}} & \multicolumn{2}{c}{OpenImages-} \\
                                                                    & Common        & Uncommon        \\ \hline
BCE                                                                 & 81.1          & 65.4            \\
CE                                                                  & 83.1          & 67.7            \\
Hill                                                                & 82.7          & 69.2            \\
ASL                                                                 & 83.2          & 70.2            \\ \shline
\end{tabular}
\caption{\textbf{Performance comparison of different alignment loss functions for image-text alignment.}}
\label{tab:ita-loss}
\end{table}

% \usepackage{multirow}
\begin{table}[h]
\footnotesize
\centering
\begin{tabular}{c|cc}
\shline
\multirow{2}{*}{\begin{tabular}[c]{@{}c@{}}Tagging\\ Loss\end{tabular}} & \multicolumn{2}{c}{OpenImages-} \\
                                                                    & Common        & Uncommon        \\ \hline
Hill                                                                  & 79.6          & 67.7            \\
SPLC                                                                & 82.0          & 66.3            \\
ASL                                                                 & 83.2          & 70.2            \\ \shline
\end{tabular}
\caption{\textbf{Performance comparison of different alignment loss functions for image tagging.}}
\label{tab:tagging-loss}
\end{table}

\section{Model Architecture Comparison}

\vspace{5pt}
\noindent
\textbf{Off-The-Shelf Text Encoder.} In this section, we explore the impact of different off-the-shelf text encoders, including pre-trained BERT~\cite{devlin2018bert} and CLIP text encoder. Table~\ref{tab:text-encoder} showcases that the text/tag embedding extracted by CLIP text encoder is much better than that extracted by BERT. This suggest the image aligned text features can effectively enhance the ability of image text alignment models, especially when the text encoder remains frozen.

% \usepackage{multirow}
\begin{table}[h]
\footnotesize
\centering
\begin{tabular}{c|cc}
\shline
\multirow{2}{*}{\begin{tabular}[c]{@{}c@{}}Text\\ Encoder\end{tabular}} & \multicolumn{2}{c}{ImageNet-} \\
                                                                        & Common       & Uncommon      \\ \hline
BERT                                                                    & 57.9         & 24.2          \\
CLIP                                                                    & 63.6         & 44.6          \\ \shline
\end{tabular}
\caption{\textbf{Performance comparison of different off-the-shelf text encoders.}}
\label{tab:text-encoder}
\end{table}

\begin{figure*} [t]
\centering
  \includegraphics[width=0.95\linewidth]{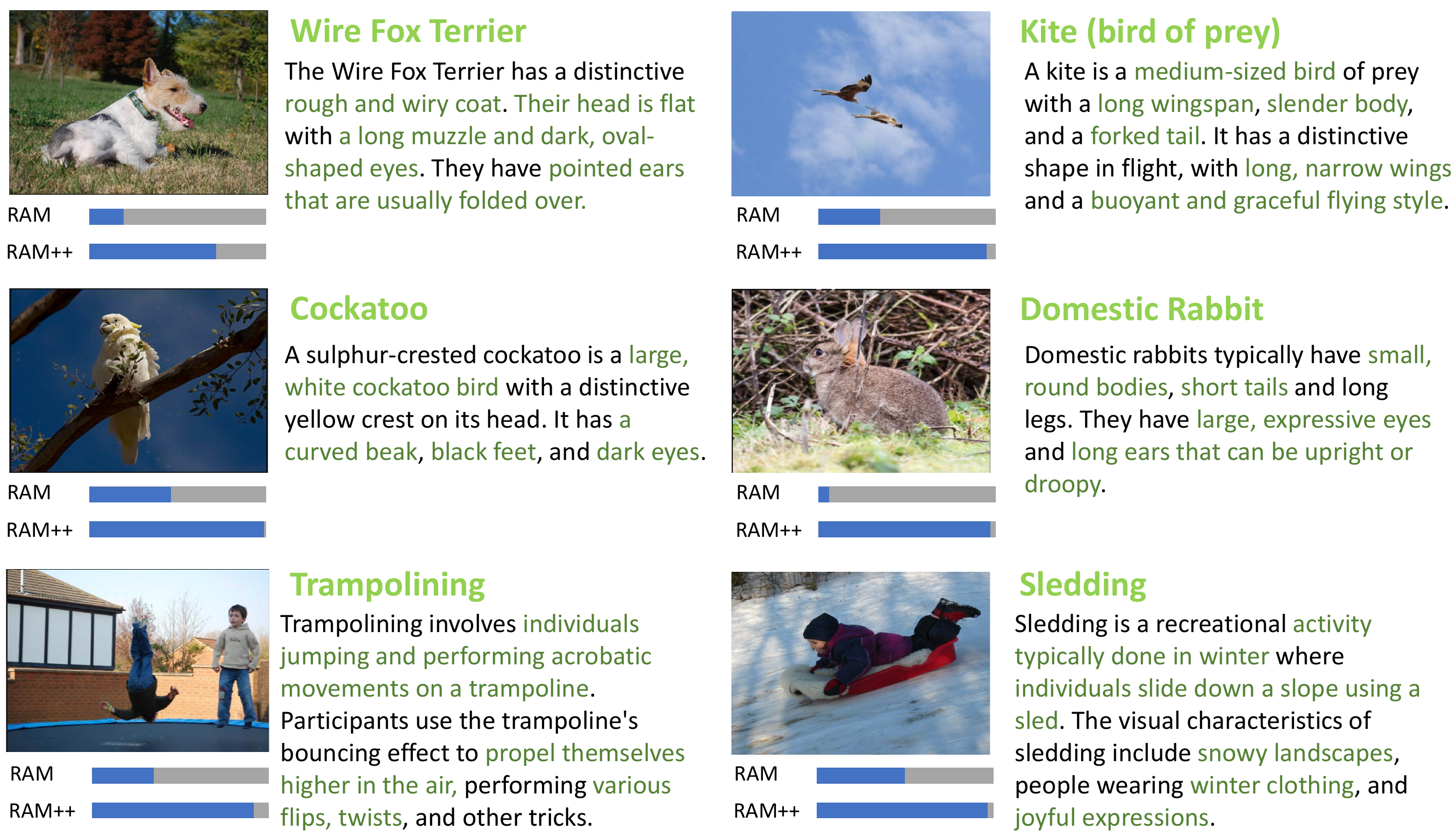}
  % \vspace{-0.8em}
  \caption{\textbf{More visual comparison of model prediction probabilities} between RAM and RAM++ for open-set category recognition. RAM++ effectively utilizes visual features derived from the descriptions, demonstrating a significant improvement on prediction probabilities.}
  \label{fig:visual-examples-appendix}
  % \vspace{-0.8em}
\end{figure*}

\vspace{5pt}
\noindent
\textbf{Larger Image Encoder.} Table~\ref{tab:image-encoder} presents the performance comparison of image encoders with different scales. While Swin$_{Large}$ exhibits improvements on predefined categories, it reveals a decrease on performance for open-set categories.

% % Please add the following required packages to your document preamble:
% % \usepackage{multirow}
% \begin{table}[h]
% \footnotesize
% \centering
% \begin{tabular}{c|cc}
% \shline
% \multirow{2}{*}{\begin{tabular}[c]{@{}c@{}}Image\\ Encoder\end{tabular}} & \multicolumn{2}{c}{ImageNet-} \\
%                                                                          & Common       & Uncommon       \\ \hline
% Swin-B                                                                   & 72.4         & 55.0           \\
% Swin-L                                                                   & 74.0         & 53.4           \\ \shline
% \end{tabular}
% \caption{Performance comparison of different image encoder.}
% \label{tab:image-encoder}
% \end{table}

% \usepackage{multirow}
\begin{table}[h]
\centering
 \footnotesize
  \resizebox{\linewidth}{!}{
\begin{tabular}{c|cc|cc|c}
\shline
\multirow{2}{*}{\begin{tabular}[c]{@{}c@{}}Image\\ Encoder\end{tabular}} & \multicolumn{2}{c|}{Tag-Common} & \multicolumn{2}{c|}{Tag-Uncommon} & Phrase \\
                                                                         & Openimages   & ImageNet   & Openimages    & ImageNet    & HICO       \\ \hline
Swin-B                                                                   & 86.6         & 72.4             & 75.4          & 55.0              & 37.7       \\
Swin-L                                                                   & 86.4         & 74.0             & 75.0          & 53.4              & 39.2       \\ \shline
\end{tabular}
}
\caption{\textbf{Performance comparison of different image encoder.}}
\label{tab:image-encoder}
\end{table}

\vspace{5pt}
\noindent
\textbf{Depth of Alignment Decoder.} Table~\ref{tab:decoder-depth} demonstrates that increasing the layer depth of the alignment decoder does not necessarily enhance the model's recognition capabilities, allowing ITA to achieve superior performance with minimal computational consumption.

% \usepackage{multirow}
\begin{table}[h]
\footnotesize
\centering
\begin{tabular}{c|cc}
\shline
\multirow{2}{*}{\begin{tabular}[c]{@{}c@{}}Decoder\\ Depth\end{tabular}} & \multicolumn{2}{c}{OpenImages-} \\
                                                                                      & Common        & Uncommon        \\ \hline
2                                                                                     & 82.4          & 61.7            \\
6                                                                                     & 80.2          & 58.5            \\ \shline
\end{tabular}
\caption{\textbf{Performance comparison of different layer depth for alignment decoder.}}
\label{tab:decoder-depth}
\end{table}

\section{Additional Qualitative Results}

In Figure~\ref{fig:visual-examples-appendix}, we show more examples that RAM++ presents better robustness on open-set categories against RAM, by utilizing visual features derived from the tag descriptions.

\section{Evaluation Benchmark Details}

\begin{figure}[]
% 	\subfigure{
    	\begin{minipage}[t]{0.5\linewidth}
    	\centering
    	\includegraphics[width=1.6in]{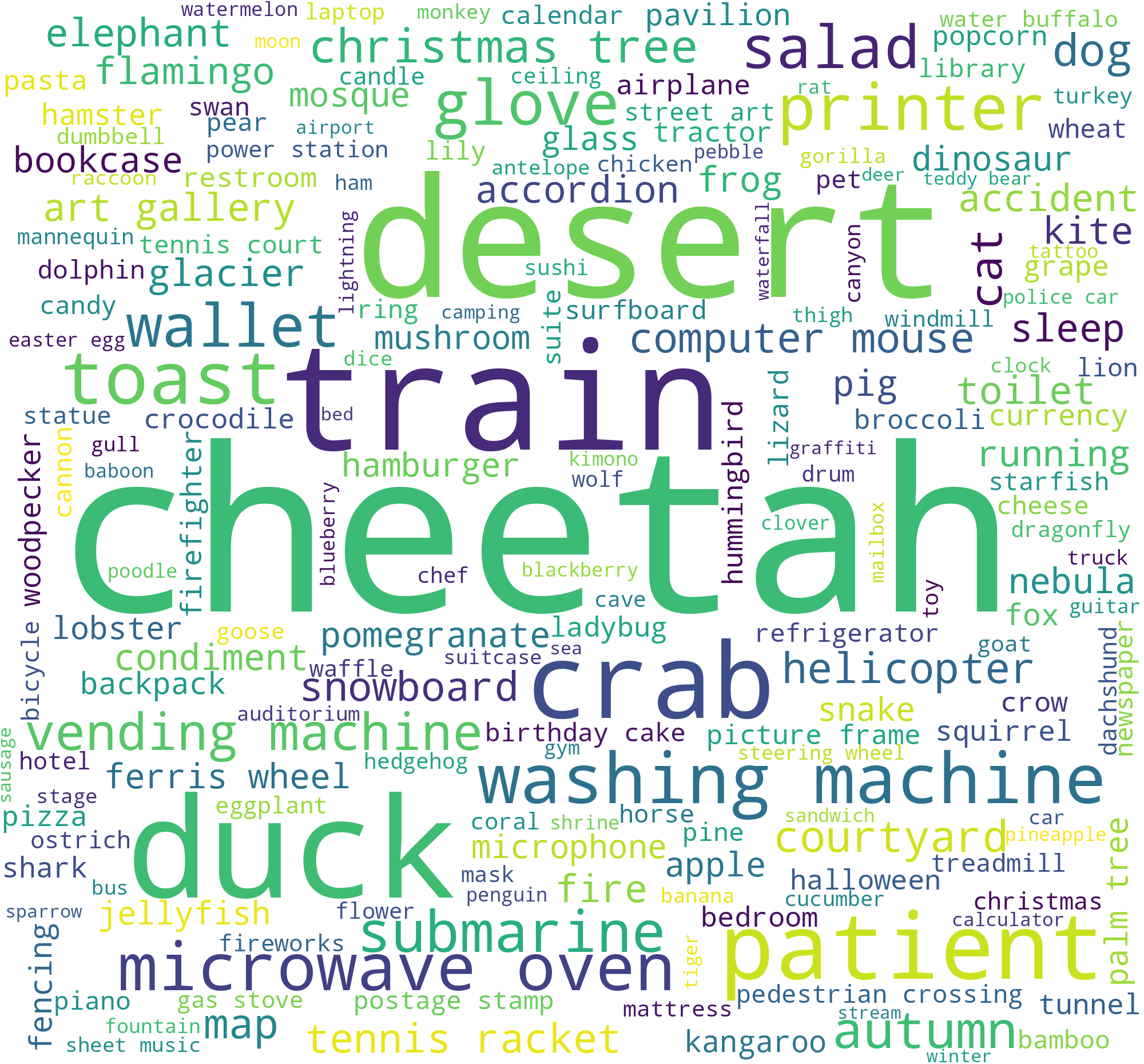}
    	\centerline{\scriptsize (a)~OpenImages-Common}
    	\centering
    	\includegraphics[width=1.6in]{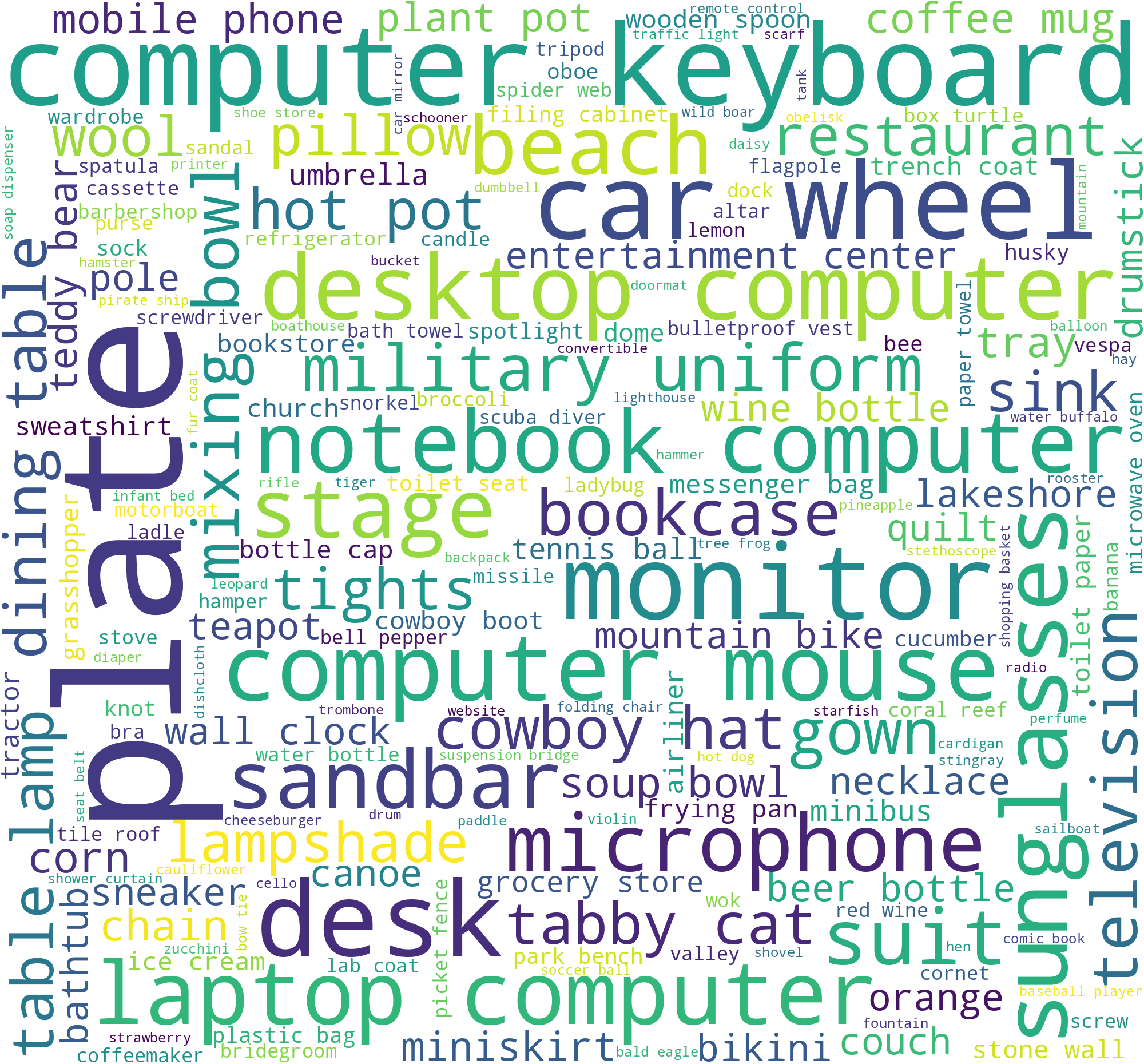}
    	\centerline{\scriptsize (c)~ImageNet-Common}
	    \end{minipage}%
% 	}%
% 	\subfigure{
    	\begin{minipage}[t]{0.5\linewidth}
    	\centering
    	\includegraphics[width=1.6in]{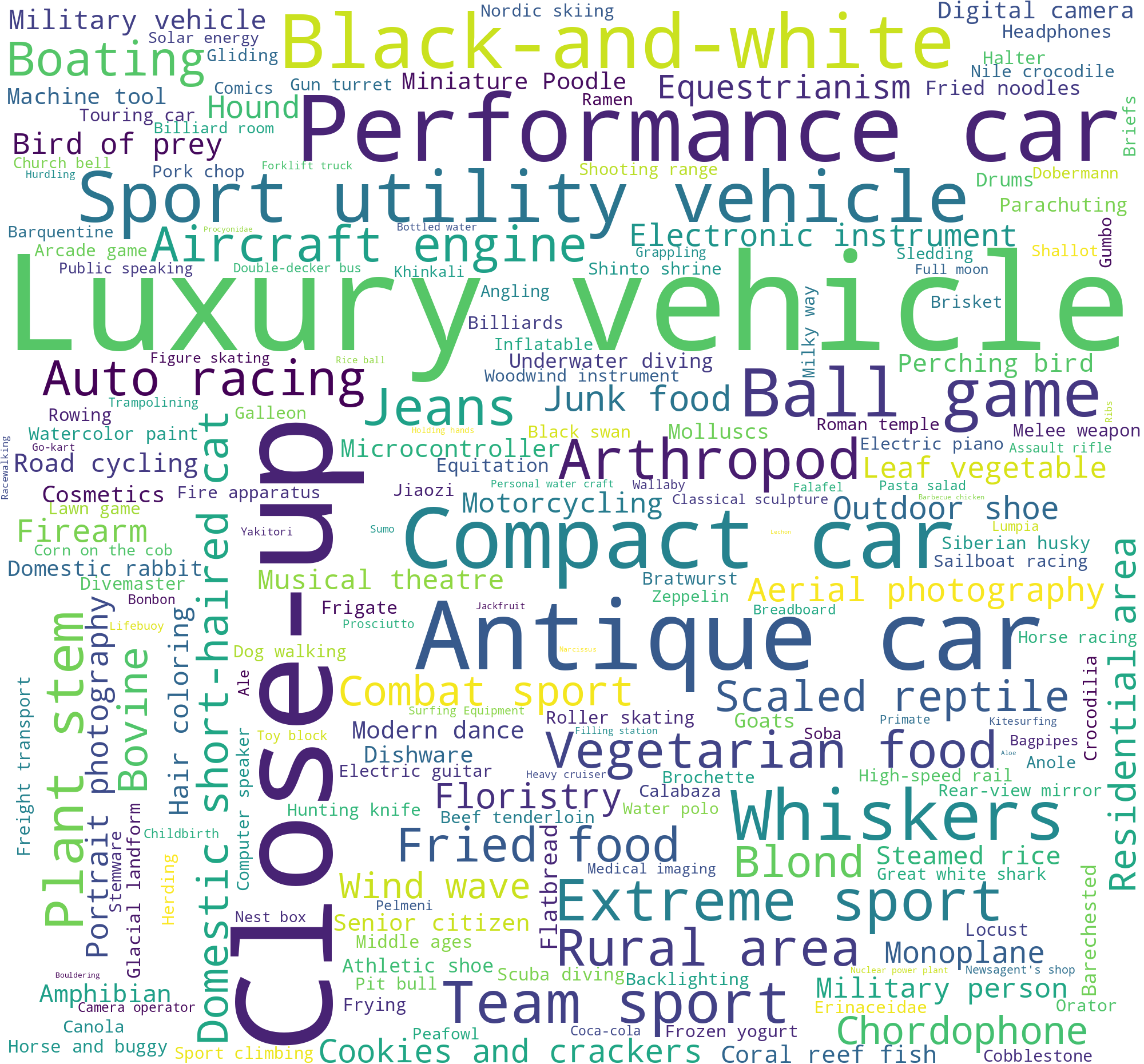}
    	\centerline{\scriptsize (b)OpenImages-Uncommon}
    	\centering
    	\includegraphics[width=1.6in]{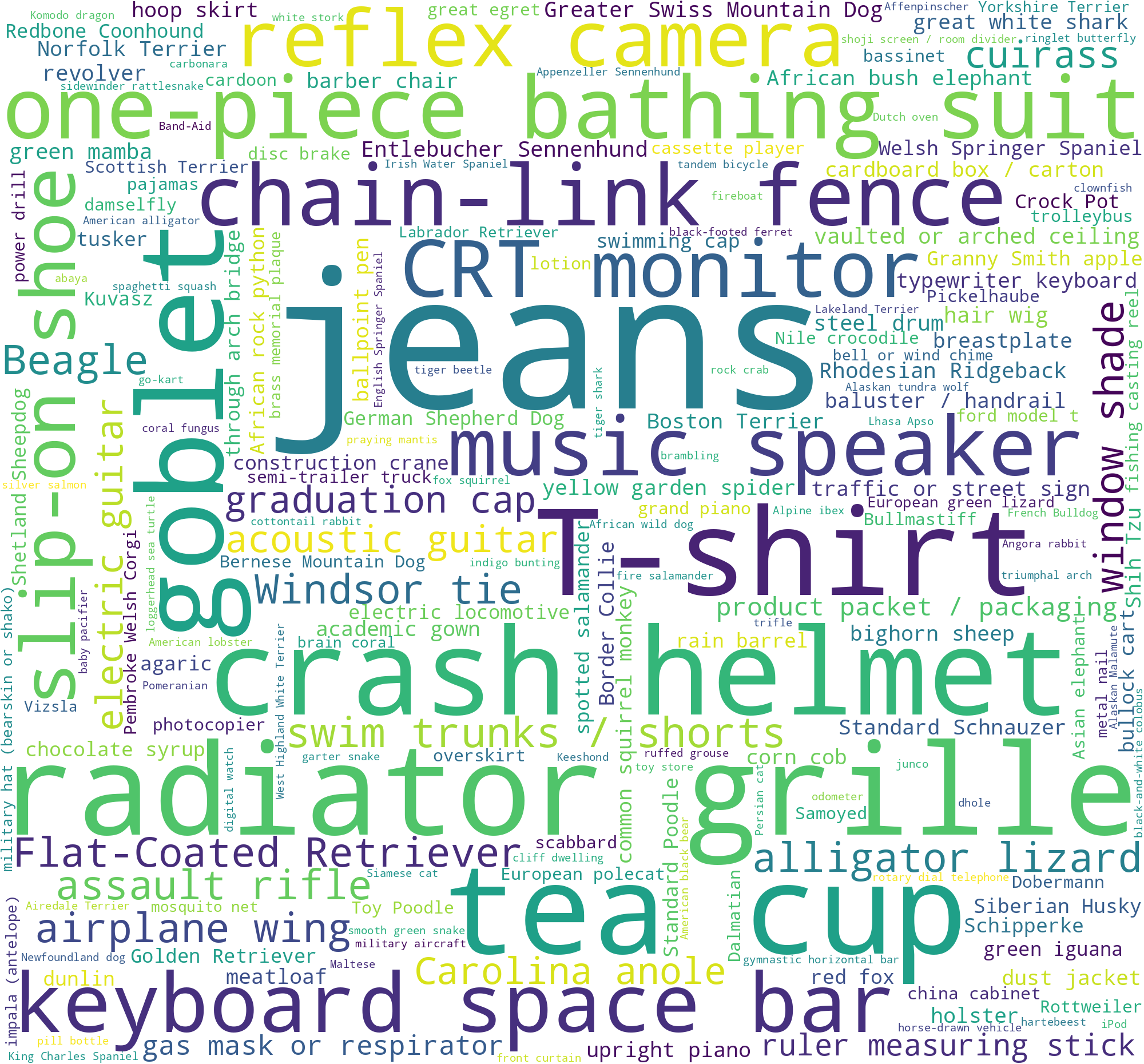}
    	\centerline{\scriptsize (d)~ImageNet-Uncommon}
    	\end{minipage}%
% 	}%
	\centering
	\caption{\textbf{Illustration of the categories in various evaluation benchmarks.}}
    \label{img:app_benchmark_categories}
\end{figure}

In Figure~\ref{img:app_benchmark_categories}, we present the word clouds of the categories in various evaluation benchmarks. The word size is proportional to the category frequency. This visualization reveals that uncommon categories not included in the predefined labeling systems are not necessarily rare categories. Instead, most of them are well-recognized and commonly understood categories.

% It can be seen that the uncommon categories which are not covered on predefined label system are not rare categories, but often well-known categories.

\end{document}